\documentclass{article}

\PassOptionsToPackage{numbers,compress}{natbib}
 \usepackage[preprint]{neurips_2026}

\usepackage[utf8]{inputenc} % allow utf-8 input
\usepackage[T1]{fontenc}    % use 8-bit T1 fonts
\usepackage[hidelinks]{hyperref}       % hyperlinks
\usepackage{url}            % simple URL typesetting
\usepackage{booktabs}       % professional-quality tables
\usepackage{amsmath}
\usepackage{amsfonts}       % blackboard math symbols
\usepackage{nicefrac}       % compact symbols for 1/2, etc.
\usepackage{microtype}      % microtypography
\usepackage[table]{xcolor}  % colors
\usepackage{graphicx}
\usepackage{tabularx}
\usepackage{array}
\usepackage{makecell}
\usepackage{multirow}
\usepackage{wrapfig}
\usepackage{xurl}
\usepackage{enumitem}
\usepackage{algorithm}
\usepackage{algpseudocode}
\usepackage{float}
\setlist[itemize]{leftmargin=1.4em, topsep=0pt, partopsep=0pt, itemsep=0pt}
\title{
    \makebox[0pt][r]{%
        $\vcenter{\hbox{\includegraphics[width=0.09\textwidth]{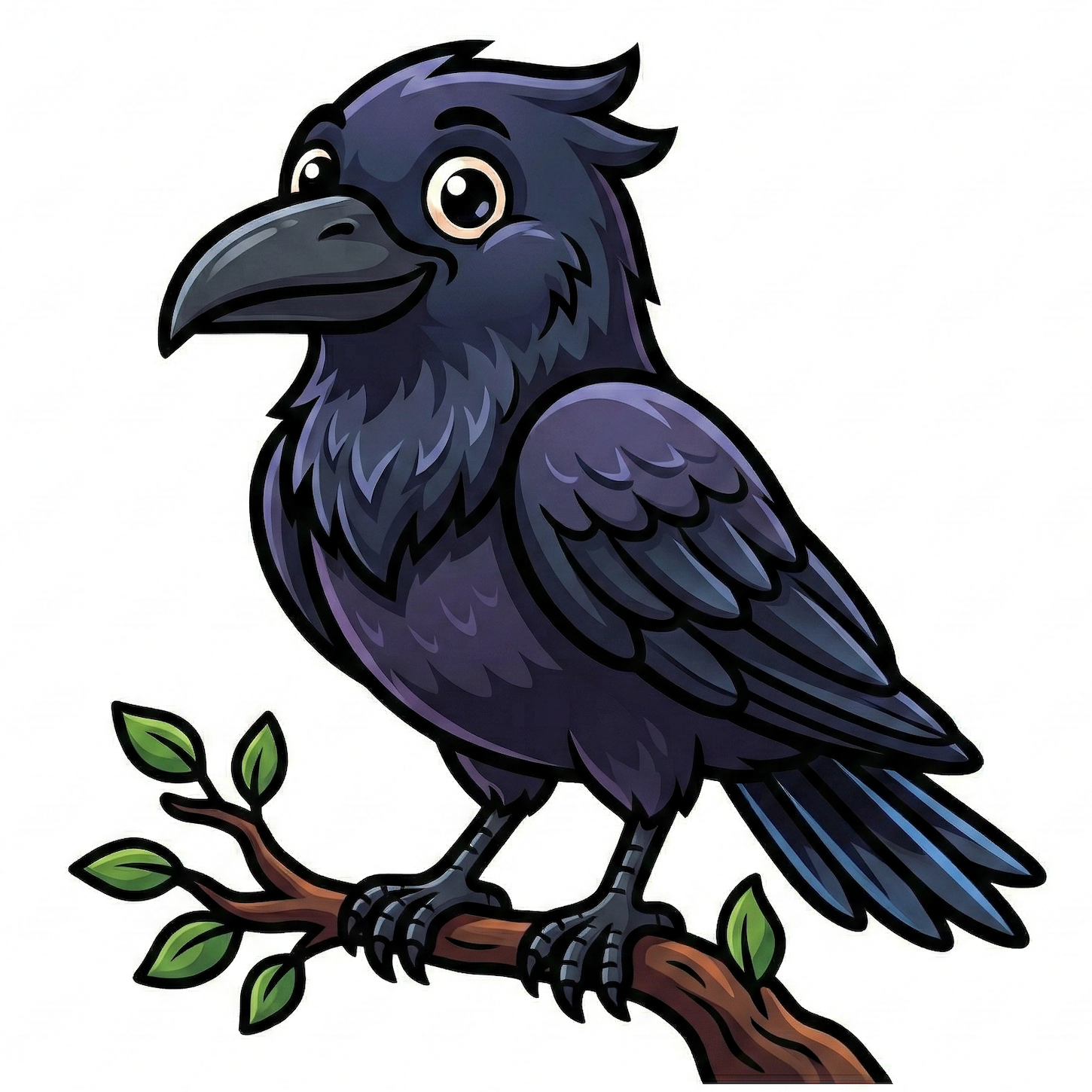}}}$
    }
    $\vcenter{\hbox{\begin{tabular}{@{}c@{}}
        RAVEN: Real-time Autoregressive Video \\
        Extrapolation with Consistency-model GRPO
    \end{tabular}}}$
}

\author{%
  Yanzuo Lu \quad Ronglai Zuo \quad Jiankang Deng \\
  Imperial College London \\
  \url{https://yanzuo.lu/raven}
}

\begin{document}

\maketitle

\begin{abstract}
Causal autoregressive video diffusion models support real-time streaming generation by extrapolating future chunks from previously generated content.
Distilling such generators from high-fidelity bidirectional teachers yields competitive few-step models, yet a persistent gap between the history distributions encountered during training and those arising at inference constrains generation quality over long horizons.
We introduce the Real-time Autoregressive Video Extrapolation Network (RAVEN), a training-time test framework that repacks each self rollout into an interleaved sequence of clean historical endpoints and noisy denoising states.
This formulation aligns training attention with inference-time extrapolation and allows downstream chunk losses to supervise the history representations on which future predictions depend.
We further propose Consistency-model Group Relative Policy Optimization (CM-GRPO), which reformulates a consistency sampling step as a conditional Gaussian transition and applies online Reinforcement Learning (RL) directly to this kernel, avoiding the Euler-Maruyama auxiliary process adopted in prior flow-model RL formulations.
Experiments demonstrate that RAVEN surpasses recent causal video distillation baselines across quality, semantic, and dynamic degree evaluations, and that CM-GRPO provides further gains when combined with RAVEN.
\end{abstract}

\section{Introduction}

Recent progress in video diffusion has established bidirectional models as the dominant paradigm for high-fidelity generation~\cite{arkhipkin2025kandinsky, chen2026sanavideo, gao2025seedance, genmo2024mochi, hacohen2024ltxvideob, hacohen2026ltx2, kong2025hunyuanvideo, ma2025stepvideot2v, meta2024movie, nvidia2025cosmos, seedance2025seedance, seedance2026seedance, wan2025wan, wu2025hunyuanvideo, yang2024cogvideox}.
Their reliance on bidirectional context and a large number of denoising steps, however, limits their suitability for real-time generation, where video must be produced continuously as a stream.
This requirement has motivated causal autoregressive architectures that extrapolate future chunks from previously generated content~\cite{ai2025magi1a, bai2026causality, chen2025skyreelsv2a, deng2025autoregressive, guo2025endtoend, huang2026live, jin2024pyramidal, li2025stable, liu2025infinitystar, po2025bagger, wu2025pack, xiang2025macrofrommicro, yuan2026helios, zhang2025blockvid, zhang2026pretraining}.
The strongest generation capability still largely resides in high-step bidirectional models, and recent work has studied asymmetric distillation, which transfers knowledge from such bidirectional teachers to causal student generators~\cite{huang2025selfa, liu2025rolling, lu2025reward, yang2025longlivea, yin2025slowa, zhu2026causal}.
The resulting few-step generators achieve real-time generation speeds while retaining much of the visual fidelity of their teachers.

\begin{figure}[t]
    \centering
    \includegraphics[width=0.975\linewidth]{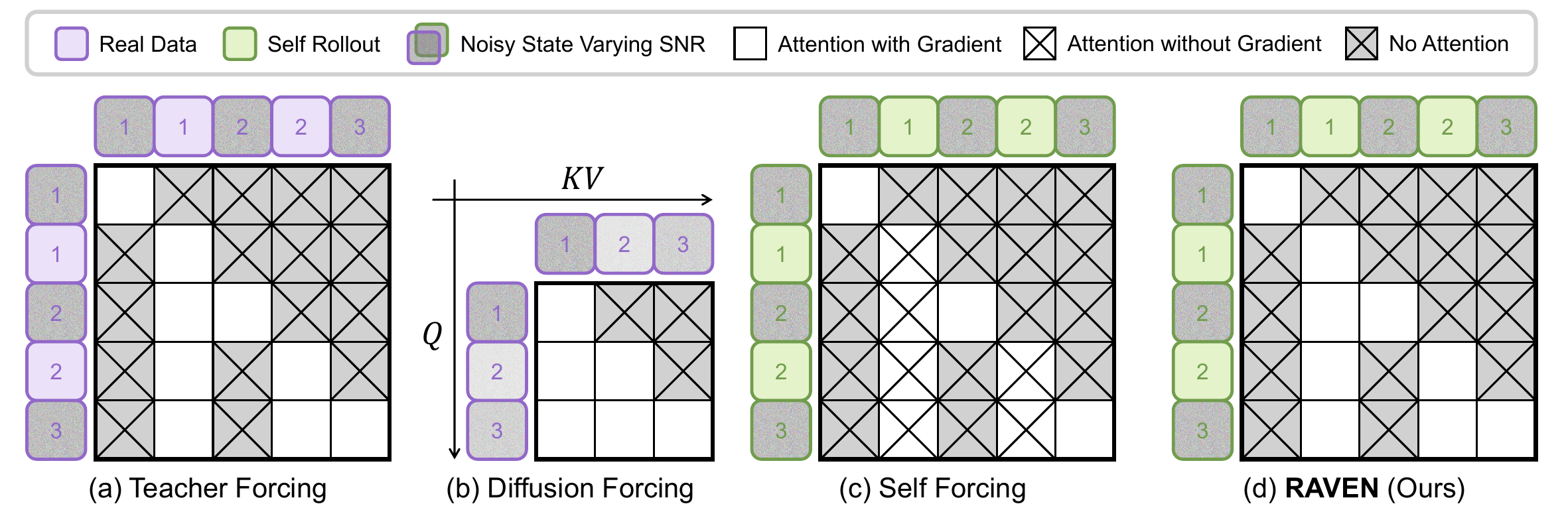}
    \vspace{-7.5pt}
    \caption{\textbf{Attention Mask Configuration.}
    Autoregressive video diffusion training paradigms differ in how historical states enter attention and whether those states receive end-to-end supervision from later chunks.
    Teacher Forcing and Diffusion Forcing rely on data-driven historical states, inducing a training distribution that differs from inference.
    Self Forcing shifts the history distribution toward inference but reuses the historical cache as detached context.
    RAVEN instead repacks each self rollout into clean historical endpoints and noisy denoising states, allowing later chunks to attend to the same history used during extrapolation while their losses supervise the cached representations.}
    \vspace{-5pt}
    \label{fig:rollout}
\end{figure}

A central challenge in autoregressive video diffusion distillation lies in how the model represents and reuses historical chunks, as each generated chunk becomes the context on which all subsequent predictions depend.
As illustrated in Figure~\ref{fig:rollout}, existing training paradigms differ in both the source of historical states and whether those states receive end-to-end supervision from later chunks.
Teacher Forcing trains with real historical chunks, which provides clean supervision but does not expose the generator to its own test-time history.
Diffusion Forcing~\cite{chen2024diffusiona,song2025historyguided} trains causal diffusion models by assigning each token an independently sampled Signal-to-Noise Ratio (SNR), and CausVid~\cite{yin2025slowa} adapts this construction to autoregressive video distillation by incorporating Distribution Matching Distillation (DMD)~\cite{yin2024onestep,yin2024improved}.
This formulation optimizes the causal generator under a history distribution that does not match inference, and the resulting discrepancy can accumulate across autoregressive rollouts.
Self Forcing~\cite{huang2025selfa} reduces this discrepancy by conditioning the DMD objective on self rollouts, yet the historical cache is reused as detached context, so the history representations receive no end-to-end supervision from subsequent chunk losses.

We propose the Real-time Autoregressive Video Extrapolation Network (RAVEN), a training-time test framework that directly supervises the history construction used during autoregressive extrapolation.
Starting from self rollouts of the few-step causal generator, RAVEN repacks the sampled trajectory into an interleaved sequence of clean historical endpoints and noisy denoising states.
Within this sequence, clean rollout chunks provide the causal history for subsequent predictions, while noisy states from the same rollout remain the supervised denoising inputs.
The resulting attention computation aligns more closely with inference than Teacher Forcing or Diffusion Forcing and keeps history representations inside the supervised forward pass, as shown in Figure~\ref{fig:rollout}(d).
This design enables gradients from later chunks to shape the cached representations on which future predictions depend, while avoiding the cost of backpropagating through an entire autoregressive sampling trajectory.

Reinforcement learning (RL) has become an influential post-training paradigm for large generative models, and recent work has begun to adapt it to diffusion and flow models.
Flow-GRPO~\cite{liu2025flowgrpo} demonstrates this direction for flow matching, addressing the conflict between deterministic Ordinary Differential Equation (ODE) sampling and the stochastic exploration required by policy optimization through an ODE-to-Stochastic Differential Equation (SDE) conversion followed by Euler-Maruyama discretization.
The causal generator in RAVEN employs a few-step consistency sampler, for which Euler-Maruyama introduces a train-test discrepancy by optimizing over stochastic transitions that differ from the deterministic sampling used at inference.
We observe that a consistency sampling step can be cast as a conditional Gaussian transition parameterized by the predicted clean endpoint, enabling the policy objective to be defined on the same update rule used during generation without an auxiliary stochastic process.
This correspondence is especially consequential for autoregressive video generation, where each generated chunk alters the history on which subsequent predictions depend.
We therefore propose Consistency-model Group Relative Policy Optimization (CM-GRPO), which applies group relative policy optimization directly to this consistency transition kernel.

Our contributions are as follows.
\begin{itemize}
    \item We identify a history supervision gap in autoregressive video diffusion distillation, where existing methods are either optimized under history distributions that differ from inference or conditioned on rollout history without end-to-end supervision.
    \item We introduce RAVEN, a training-time test framework that repacks self rollouts into an interleaved sequence of clean historical endpoints and noisy denoising states, allowing supervision to propagate through the history representations used during extrapolation.
    \item We propose CM-GRPO, which reformulates a consistency sampling step as a conditional Gaussian transition kernel and applies group relative policy optimization directly to this kernel, matching the sampler interface used at inference.
    \item We demonstrate that RAVEN surpasses recent causal video distillation baselines and that CM-GRPO provides complementary gains when combined with RAVEN.
\end{itemize}

\section{Related Work}

\noindent\textbf{Autoregressive Video Diffusion Distillation.}
Autoregressive video generation encompasses several parallel directions beyond the causal distillation setting studied in this paper.
One line of work explores the design of the autoregressive rollout itself, either extending the prediction window for longer sequences or conditioning on intermediate noisy latents rather than fully denoised outputs as historical context~\cite{chen2026context, cui2025selfforcinga, liu2025rolling, liu2026streaming, yang2025longlivea, zou2026hiar}.
Although our current implementation conditions on clean latents, the training-time test paradigm can simulate these alternative history mechanisms to provide end-to-end supervision.
A separate direction develops architectures with dedicated temporal memory for managing long-range context during training~\cite{chen2026context, chen2026futureinformed, ji2025memflow, shin2026motionstream, yu2025videossm, zhu2025memorizeandgenerate}, while a complementary body of training-free methods adapts models at inference time for length extrapolation~\cite{cui2026lol, li2026train, xiang2026pathwise, yesiltepe2026infinityrope, yi2025deep, zhao2026relax}.
Our framework is orthogonal to both families, as any strategy that generates and caches the next chunk through specialized memory designs can be executed within the self-rollout phase and benefit from the subsequent interleaved optimization.

\noindent\textbf{Online RL in Diffusion Model.}
Online RL has become a practical paradigm for aligning diffusion and flow models after pretraining, beginning with reward-guided optimization for image generation and gradually evolving into policy optimization methods tailored to diffusion and flow trajectories~\cite{black2024training, liu2025flowgrpo, wang2026prefgrpo, xu2023imagerewarda, zheng2026diffusionnft}.
This approach has since been extended to autoregressive generators and world models, where reinforcement learning serves not only for preference alignment but also for preserving pretrained capabilities and improving controllable generation over long horizons~\cite{ma2025stage, wang2026worldcompass, wu2025rlvrworld, ye2025reinforcement, zhang2026astrolabe, zhao2026realtime}.
Parallel work applies online RL to distilled and few-step generators, where the central challenge is to improve alignment without sacrificing the efficiency that makes these models practical~\cite{chen2025flashdmd, guo2026erudiff, luo2026tdmr1}.
Much of the follow-up work has focused on refining the policy objective itself.
Some methods revisit regularization to control reward hacking and distribution drift~\cite{he2025gardo, liu2026unigrpo, ye2025dataregularized, zhu2026diffusion}, while others study how the stochasticity or numerical form of the sampler shapes policy optimization~\cite{he2026neighbor, luo2025reinforcing, sheng2025understanding, wang2025coefficientspreserving, wangpcflow, zhang2026egrpo, zheng2026manifoldaware}.
A separate direction makes more deliberate use of the denoising trajectory, for instance through branching, tree search, or stepwise credit assignment~\cite{chen2026superflow, ding2025treegrpo, fu2025dynamictreerpo, he2025tempflowgrpo, li2025branchgrpo, liang2026leapalign, luo2025sample, lyu2025multigrpo, mcallister2026finite, savani2026stepwise, shao2025anchoring, tang2025tr2d2, tong2026alleviating, wang2025grpoguard, yue2026know, zhang2026opgrpo, zhou2025finegrained}.
Our method is most closely related to the literature on few-step generation and sampler design.
Rather than adopting the Euler-Maruyama discretization used in prior online RL formulations for flow models, CM-GRPO formulates the policy objective directly on the consistency transition kernel and combines it with the training-time test framework of RAVEN, more closely matching the inference-time behavior of autoregressive video extrapolation.

\section{Methodology}

\subsection{Preliminaries}

Let $x_{1:T}$ denote a sequence of latent video chunks and $c$ the text condition, with hats used for student-generated quantities.
Throughout the paper, the subscript $t$ indexes the chunk position, while a superscript in parentheses, such as $(n)$, $(u)$, or $(s)$, denotes the noise level.
We write the autoregressive video diffusion model as
\begin{equation}
    p_{\theta}(x_{1:T}\mid c)=\prod_{t=1}^{T}p_{\theta}(x_t\mid h_t,c), \qquad h_t=\mathcal{H}(x_{<t}).
\end{equation}
The operator $\mathcal{H}(\cdot)$ denotes the history representation encoded by the model via its cache.
For a noise level $n$, we define the noisy current chunk as $z_t^{(n)}=\alpha_n x_t+\sigma_n \epsilon$, with $\epsilon\sim\mathcal{N}(0,I)$.
Training paradigms are distinguished primarily by how the history $h_t$ is constructed from past chunks, and we detail this distinction in the following subsections.

\noindent\textbf{History Formulation in Diffusion Forcing and Self Forcing.}
Recent methods for autoregressive video diffusion distillation are largely built on either Diffusion Forcing~\cite{chen2024diffusiona} or Self Forcing~\cite{huang2025selfa}.
In CausVid~\cite{yin2025slowa}, training follows Diffusion Forcing and represents the history as $h_t^{\mathrm{DF}}=\mathcal{H}\bigl(z_1^{(n_1)},\ldots,z_{t-1}^{(n_{t-1})}\bigr)$, perturbing each ground-truth prefix chunk with an independently sampled noise level before entering the causal context.
Self Forcing~\cite{huang2025selfa} instead unrolls the autoregressive generator at training time and reuses detached cache representations written as $h_t^{\mathrm{SF}}=\operatorname{sg}\!\left(\mathcal{H}(\hat{x}_{<t})\right)$, where the stop-gradient operator $\operatorname{sg}(\cdot)$ treats historical chunks as fixed context for subsequent denoising steps.
Both formulations therefore leave the cache construction outside end-to-end supervision, motivating the training-time test formulation introduced next.

\noindent\textbf{Euler-Maruyama Discretization in Flow-GRPO.}
Flow-GRPO~\cite{liu2025flowgrpo} starts from the rectified-flow ODE $\mathrm{d}y_{\tau}=v_{\theta}(y_{\tau},\tau,c)\,\mathrm{d}\tau$, where $y_{\tau}$ denotes the latent variable at denoising time $\tau\in[0,1]$.
To inject the stochasticity required for policy optimization, it introduces an ODE-to-SDE conversion and operates on the reverse-time SDE $\mathrm{d}y_{\tau}=b_{\theta}(y_{\tau},\tau,c)\,\mathrm{d}\tau+\sigma_{\tau}\,\mathrm{d}w$, where $b_{\theta}(y_{\tau},\tau,c)$ is the drift term and $\sigma_{\tau}\,\mathrm{d}w$ the diffusion term.
The drift term is given by
\begin{equation}
    b_{\theta}(y_{\tau},\tau,c)
    =
    v_{\theta}(y_{\tau},\tau,c)
    +\frac{\sigma_{\tau}^{2}}{2\tau}\bigl(y_{\tau}+(1-\tau)v_{\theta}(y_{\tau},\tau,c)\bigr).
\end{equation}
Applying Euler-Maruyama discretization yields
\begin{equation}
    y_{\tau+\Delta\tau}
    =
    y_{\tau}
    +b_{\theta}(y_{\tau},\tau,c)\Delta\tau
    +\sigma_{\tau}\sqrt{\Delta\tau}\,\epsilon,
    \quad \epsilon\sim\mathcal{N}(0,I).
\end{equation}
Equivalently, the Euler-Maruyama step defines an isotropic Gaussian policy kernel,
\begin{equation}
    \pi_{\theta}^{\mathrm{EM}}\bigl(y_{\tau+\Delta\tau}\mid y_{\tau},c\bigr)
    =
    \mathcal{N}\!\left(
        y_{\tau+\Delta\tau};
        y_{\tau}+b_{\theta}(y_{\tau},\tau,c)\Delta\tau,
        \sigma_{\tau}^{2}\Delta\tau I
    \right).
\end{equation}
This auxiliary kernel makes the policy ratio and the KL term tractable in closed form, but its stochastic transitions remain absent from the deterministic ODE sampler used at inference.
ODE-based samplers are typically deterministic~\cite{lin2024sdxllightning, lu2025hyperbagel, luadversarial, ren2024hypersd, salimans2022progressive, wang2024phased, yan2024perflow}, while the consistency sampler~\cite{kim2024consistency, luo2023latent, luo2023lcmlora, luo2024onestep, song2023consistency, yin2024improved, zheng2024trajectory, zhou2024adversarial, zhou2024score} is a notable exception in the few-step regime, remaining defined on the probability flow ODE trajectory while still yielding stochastic transitions that can serve as the policy interface directly.

\begin{figure}[t]
    \centering
    \includegraphics[width=\linewidth]{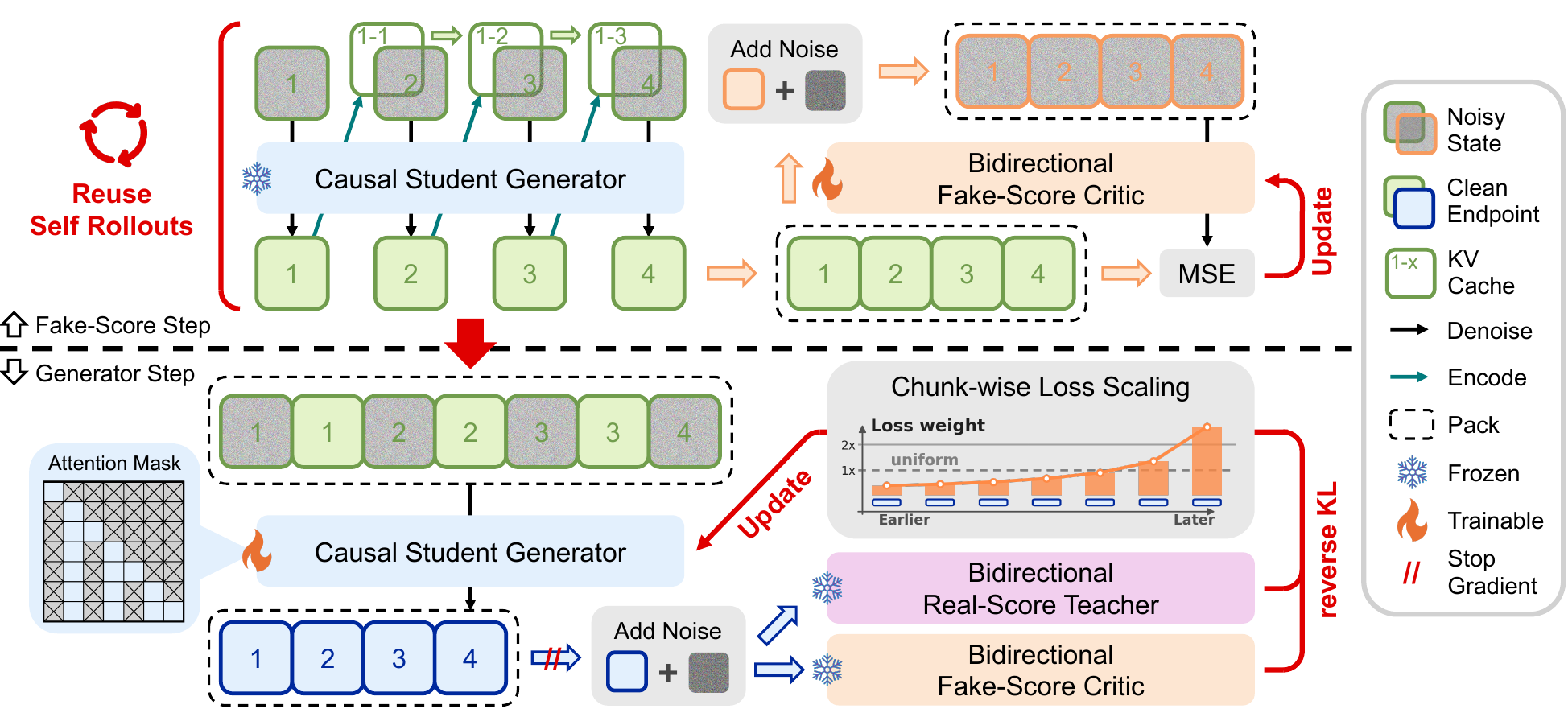}
    \vspace{-15pt}
    \caption{\textbf{Training Pipeline.}
    RAVEN builds on score distillation with a training-time test formulation that aligns the generator's training context with inference.
    In the fake-score step, the frozen generator performs autoregressive self rollout with KV cache reuse, producing the clean endpoints and noisy denoising states that are subsequently reused in the generator step.
    Rather than discarding these rollout states after critic training, RAVEN repacks them into an interleaved sequence of clean historical endpoints and noisy denoising states, processed under a causal attention mask so that each noisy state attends to the clean history the generator itself produced.
    This allows later chunk losses, scaled chunk-wise, to supervise the history representations on which future predictions depend.}
    \vspace{-5pt}
    \label{fig:pipeline}
\end{figure}

\subsection{Training-Time Test via RAVEN}

RAVEN is a training-time test framework for autoregressive video diffusion that aligns the training procedure with inference-time extrapolation.
Building upon the asymmetric distillation formulated by CausVid~\cite{yin2025slowa}, the pipeline distills knowledge from a frozen bidirectional teacher into the causal student generator.
As illustrated in Figure~\ref{fig:pipeline}, training alternates between a fake-score step and a generator step.
In the fake-score step, the bidirectional fake-score critic is updated on self-rollout samples perturbed with Gaussian noise.
In the generator step, the causal student generator is updated via a reverse Kullback-Leibler (KL) score gradient computed from evaluations by both the bidirectional real-score teacher and the learned fake-score critic.

Let $\tau_1>\cdots>\tau_K=0$ denote the few-step sampling timesteps of the consistency sampler adopted by the generator.
During the fake-score step, the frozen causal student generator autoregressively produces, for each chunk index $t$, a full denoising trajectory $\{\hat{z}_t^{(\tau_k)}\}_{k=1}^{K}$ along with the clean endpoint $\hat{x}_t=\hat{z}_t^{(0)}$.
These clean endpoints are perturbed with Gaussian noise to form the training inputs for the fake-score critic.
During the generator step, the same self rollout is reused and the noisy state at denoising level $\tau_k$ is taken directly from each chunk's sampled trajectory.
These rollout states are then packed into an input sequence processed under the attention mask illustrated in Figure~\ref{fig:rollout}(d).
Specifically, for a sampled timestep $u\in\{\tau_1,\ldots,\tau_{K-1}\}$, the interleaved sequence takes the form
\begin{equation}
    \label{eq:interleaved}
    \mathcal{I}_{u}=
    \bigl(
        \hat{z}_{1}^{(u)}, \hat{x}_{1},
        \hat{z}_{2}^{(u)}, \hat{x}_{2},
        \ldots,
        \hat{z}_{T-1}^{(u)}, \hat{x}_{T-1},
        \hat{z}_{T}^{(u)}
    \bigr),
\end{equation}
where $\hat{z}_{t}^{(u)}$ is the noisy state of chunk $t$ at denoising level $u$ and $\hat{x}_{t}$ is the corresponding clean endpoint.
Within this sequence, the noisy states serve as supervised denoising targets, while the clean endpoints preceding chunk $t$ constitute its history $h_t^{\mathrm{RAVEN}}=\mathcal{H}(\hat{x}_{<t})$.
The causal student generator encodes these clean endpoints as history representations within the same forward pass, allowing later noisy states to attend to them under the causal attention structure employed during autoregressive extrapolation.
The resulting predictions are subsequently perturbed with Gaussian noise and evaluated by the bidirectional real-score teacher and the fake-score critic to compute the reverse KL score gradient.

% \noindent\textbf{History-Agnostic Formulation.}
% Although the current implementation uses clean chunks as historical context, the interleaved sequence construction is not restricted to this choice.
% It can accommodate arbitrary history representations derived from the rollout, including intermediate noisy states from preceding chunks, architecture-specific memory tokens, and cache management strategies such as sliding windows or attention sinks, provided that the same representation is reused by subsequent chunks during self rollout.
% All such components can be instantiated inside the supervised forward pass and optimized through downstream losses on later noisy states.
% RAVEN thus treats history handling as an end-to-end optimizable component of the rollout, rather than presupposing a fixed form of cached context.

\noindent\textbf{Reuse Self Rollouts.}
The formulation is inspired by the training-time test principle of EAGLE-3~\cite{li2025eagle3}, where the model is trained on the context it will produce and encounter during speculative decoding.
In language generation, this amounts to feeding a predicted draft token representation into the next simulated drafting step.
The analogous construction is substantially more involved for autoregressive video diffusion, since each chunk is the endpoint of a multi-step denoising trajectory and future chunks depend on the resulting cache.
A direct simulation would require unrolling the generator across all chunks and denoising steps within a single computation graph, incurring backpropagation through both autoregressive recursion and sampler dynamics.
RAVEN avoids this cost by exploiting the self rollout already produced during the fake-score step, which is precisely the process that defines future context at inference.
Repacking its states into an interleaved sequence, where generated clean chunks supply context and later noisy states remain supervised targets, reduces training-time test to a reorganization of existing self rollouts rather than an additional mechanism layered on top of score distillation, while faithfully preserving the dependency structure of autoregressive extrapolation.

\noindent\textbf{Chunk-wise Loss Scaling.}
Within the interleaved training sequence, chunks along the autoregressive horizon are exposed to qualitatively different denoising conditions.
Earlier chunks operate under limited historical context, whereas later chunks condition on richer accumulated history and must simultaneously maintain contextual consistency and suppress error propagation.
To account for this positional asymmetry, we introduce a future participation score.
For a sequence of $J$ chunks, let $m_j$ denote the number of scalar elements in chunk $j$ and let $\ell_j$ denote its summed loss.
The future participation score is defined as $p_j=\sum_{k=j}^{J}m_k\,/\,\sum_{k=1}^{J}m_k$, namely the fraction of supervised elements contributed by chunk $j$ and all subsequent chunks, which is larger for earlier chunks and decreases monotonically toward later ones.
The resulting profile $\mathcal{P}=(p_1,\ldots,p_J)$ is passed to a predefined weighting function $g_{\eta}$ to produce nonnegative raw weights $\tilde{w}_{1:J}=g_{\eta}(\mathcal{P})$, whose specific form is examined in the ablation studies.
For any choice of $g_{\eta}$, the normalized per-chunk weights and the aggregate chunk loss are given by $w_j=\tilde{w}_j\sum_{k=1}^{J}m_k\,/\,\sum_{k=1}^{J}\tilde{w}_k m_k$ and $\mathcal{L}_{\mathrm{chunk}}=\sum_{j=1}^{J}w_j\ell_j\,/\,\sum_{j=1}^{J}w_jm_j$.
The normalization ensures that the average element-wise weight is preserved, so $g_{\eta}$ governs only the relative distribution of gradient emphasis across chunk positions.
The complete training procedure is summarized in Algorithm~\ref{alg:raven} of Appendix~\ref{app:algorithms}.

\subsection{Online RL via CM-GRPO}

CM-GRPO is an online policy optimization method for few-step consistency generators.
As discussed in the preliminaries, Flow-GRPO~\cite{liu2025flowgrpo} achieves tractable policy optimization for flow matching by converting the deterministic ODE into an auxiliary SDE via Euler-Maruyama discretization, yet the resulting stochastic transitions are absent from the ODE sampler used at inference.
A consistency sampler, by contrast, inherently yields stochastic Gaussian transitions through its predicted clean endpoint, enabling CM-GRPO to formulate the policy objective directly on the consistency transition kernel without introducing any auxiliary stochastic process.

Consider a single consistency sampling step from noise level $u$ to a lower level $s$.
Given the current latent $\tilde{z}^{(u)}$ and condition $c$, the model predicts a clean endpoint $\hat{x}_{\theta}=f_{\theta}(\tilde{z}^{(u)},u,c)$, from which the next latent is drawn as $\tilde{z}^{(s)}=\alpha_s\hat{x}_{\theta}+\sigma_s\epsilon$ with $\epsilon\sim\mathcal{N}(0,I)$, where $\alpha_s$ and $\sigma_s$ are the noise schedule coefficients.
This sampling rule induces the Gaussian transition probability
\begin{equation}
    \label{eq:consistency_kernel}
    \pi_{\theta}\bigl(\tilde{z}^{(s)}\mid \tilde{z}^{(u)},c\bigr)
    =
    \mathcal{N}\!\left(
        \tilde{z}^{(s)};
        \mu_{\theta}^{u\to s},
        \sigma_s^2 I
    \right),
    \qquad
    \mu_{\theta}^{u\to s}=\alpha_s\hat{x}_{\theta},
\end{equation}
which constitutes the policy interface in CM-GRPO.

To instantiate group relative policy optimization on this kernel, for each condition $c$ the generator runs $G$ independent consistency trajectories, each terminating in a clean output $\hat{x}^{i}$ on which a scalar reward $R_i$ is evaluated.
Following GRPO~\cite{shao2024deepseekmath}, the group-normalized advantage is computed as $\hat{A}_i=\bigl(R_i-\operatorname{mean}(\{R_j\}_{j=1}^{G})\bigr)\,/\,\bigl(\operatorname{std}(\{R_j\}_{j=1}^{G})+\epsilon\bigr)$.
This advantage is broadcast to all consistency sampling transitions within the same trajectory, converting the endpoint reward into a per-transition objective.
For a transition from $u$ to $s$, dropping the Gaussian normalization constant and terms independent of $\theta$, the log probability under the consistency kernel reduces to
\begin{equation}
    \log \pi_{\theta}\bigl(\tilde{z}^{(s)}_i\mid \tilde{z}^{(u)}_i,c_i\bigr)
    =
    -\frac{\|\tilde{z}^{(s)}_i-\mu_{\theta}^{u\to s}\|^2}{2\sigma_s^2}.
\end{equation}
Because $\mu_{\theta}^{u\to s}=\alpha_s\hat{x}_{\theta}$, the gradient of the advantage-weighted log probability with respect to the predicted clean endpoint takes the form
\begin{equation}
    \nabla_{\hat{x}_{\theta}}
    \left[-\hat{A}_i\log \pi_{\theta}(\tilde{z}^{(s)}_i\mid \tilde{z}^{(u)}_i,c_i)\right]
    =
    -\hat{A}_i\alpha_s
    \frac{\tilde{z}^{(s)}_i-\mu_{\theta}^{u\to s}}{\sigma_s^2}.
\end{equation}
CM-GRPO implements this update through the stop-gradient regression objective
\begin{equation}
    \label{eq:cmgrpo_loss}
    \mathcal{L}_{\mathrm{CM\text{-}GRPO}}
    =
    \mathbb{E}_{i,u,s}
    \left[
        \left\|\hat{x}_{\theta}
        -\operatorname{sg}\!\left(
            \hat{x}_{\theta}
            +\frac{\hat{A}_i\alpha_s}{2\sigma_s^2}
            \bigl(\tilde{z}^{(s)}_i-\mu_{\theta}^{u\to s}\bigr)
        \right)\right\|^2
    \right],
\end{equation}
whose gradient with respect to $\hat{x}_{\theta}$ recovers exactly the endpoint gradient derived above, matching the score gradient update used in our implementation.
The same formulation also admits reference policy KL regularization.
If a reference consistency model produces a clean endpoint $\hat{x}_{\mathrm{ref}}$ under the same noisy state $\tilde{z}^{(u)}$, the KL divergence between the two Gaussian kernels reduces to
\begin{equation}
    D_{\mathrm{KL}}\!\left(
        \pi_{\theta}(\cdot\mid \tilde{z}^{(u)},c)
        \middle\|
        \pi_{\mathrm{ref}}(\cdot\mid \tilde{z}^{(u)},c)
    \right)
    =
    \frac{\alpha_s^2\|\hat{x}_{\theta}-\hat{x}_{\mathrm{ref}}\|^2}{2\sigma_s^2}.
\end{equation}
This regularizer is tractable in principle, but in our current implementation the bidirectional teacher cannot be sampled through the consistency interface and therefore does not provide $\hat{x}_{\mathrm{ref}}$ on this policy interface.
We therefore derive this closed-form expression for completeness, leaving its practical application to future work in which a compatible reference consistency model is accessible.
The complete training procedure is summarized in Algorithm~\ref{alg:cmgrpo} of Appendix~\ref{app:algorithms}.

% \noindent\textbf{Task-Agnostic Formulation.}
% Although the experiments below instantiate CM-GRPO on the autoregressive generator trained by RAVEN, the formulation is not inherently tied to autoregressive video generation.
% The same policy interface applies to any few-step generator that employs a stochastic consistency sampler, extending naturally to bidirectional models and other modalities.

\noindent\textbf{Reward Composition.}
Autoregressive video reinforcement learning requires reward signals that jointly capture motion dynamics, visual fidelity, and semantic alignment.
We empirically find that overweighting visual fidelity or semantic alignment tends to encourage static generations, whereas an overly strong motion reward degrades the remaining two aspects, making reward design challenging.
This difficulty is compounded by the limited availability of reliable holistic metrics for few-step video generation.
Reward models based on vision-language models (VLMs)~\cite{liu2025improving, wang2026unified} supply useful scalar preferences, yet their preference data are typically collected from high-step or high-quality generators, introducing a distribution shift when applied to outputs of few-step distilled models.
We therefore combine VLM-based rewards with rewards derived from representation models~\cite{ke2021musiq, laion-ai2022aestheticpredictor, li2023amt, teed2020raft}.
Each reward component is normalized within the sampled group before weighted summation, balancing reward scales and preventing any single metric from dominating the group-relative advantage.

\section{Experiments}

% \subsection{Setup}

\noindent\textbf{Implementation.}
All experiments are built upon Wan2.1-T2V-1.3B~\cite{wan2025wan} as the base model with 3 latent frames per chunk, consistent with existing baselines.
RAVEN adopts the same initialization as Causal Forcing~\cite{zhu2026causal}, which grounds the causal student in ODE distillation from an autoregressive teacher to satisfy frame-level injectivity.
The CM-GRPO stage then proceeds from the RAVEN checkpoint.
For reward composition, we integrate several representation models derived from VBench~\cite{huang2024vbenchb} consisting of complementary dimensions with a VLM-based reward trained on human feedback for \textit{Text Alignment}.
The representation rewards span both temporal dynamics (i.e., \textit{Dynamic Degree} and \textit{Motion Smoothness}) and frame-level visual quality (i.e., \textit{Aesthetic Quality} and \textit{Imaging Quality}).
More implementation details can be found at Appendix \ref{app:implementation_details}.

\noindent\textbf{Evaluation.}
We adopt VBench~\cite{huang2024vbenchb} and report the \textit{Total Score} along with the \textit{Quality Score} and \textit{Semantic Score} following the protocol shared by all compared baselines.
Because the \textit{Dynamic Degree} dimension in VBench is computed from RAFT~\cite{teed2020raft} optical-flow magnitudes and therefore picks up camera jitter and temporal drift alongside genuine motion, we additionally adopt UnifiedReward-32B~\cite{wang2026unified} to evaluate dynamic degree on the full collection of 6,220 videos generated under the VBench prompt suite, as this reward model is trained on human preference annotations and carries complementary prior knowledge from its underlying VLM.
We also conduct a user study to complement these evaluations in Appendix~\ref{app:user_study}.

\subsection{Comparison with Baselines}

\begin{figure*}[t]
    \centering
    \includegraphics[width=\linewidth]{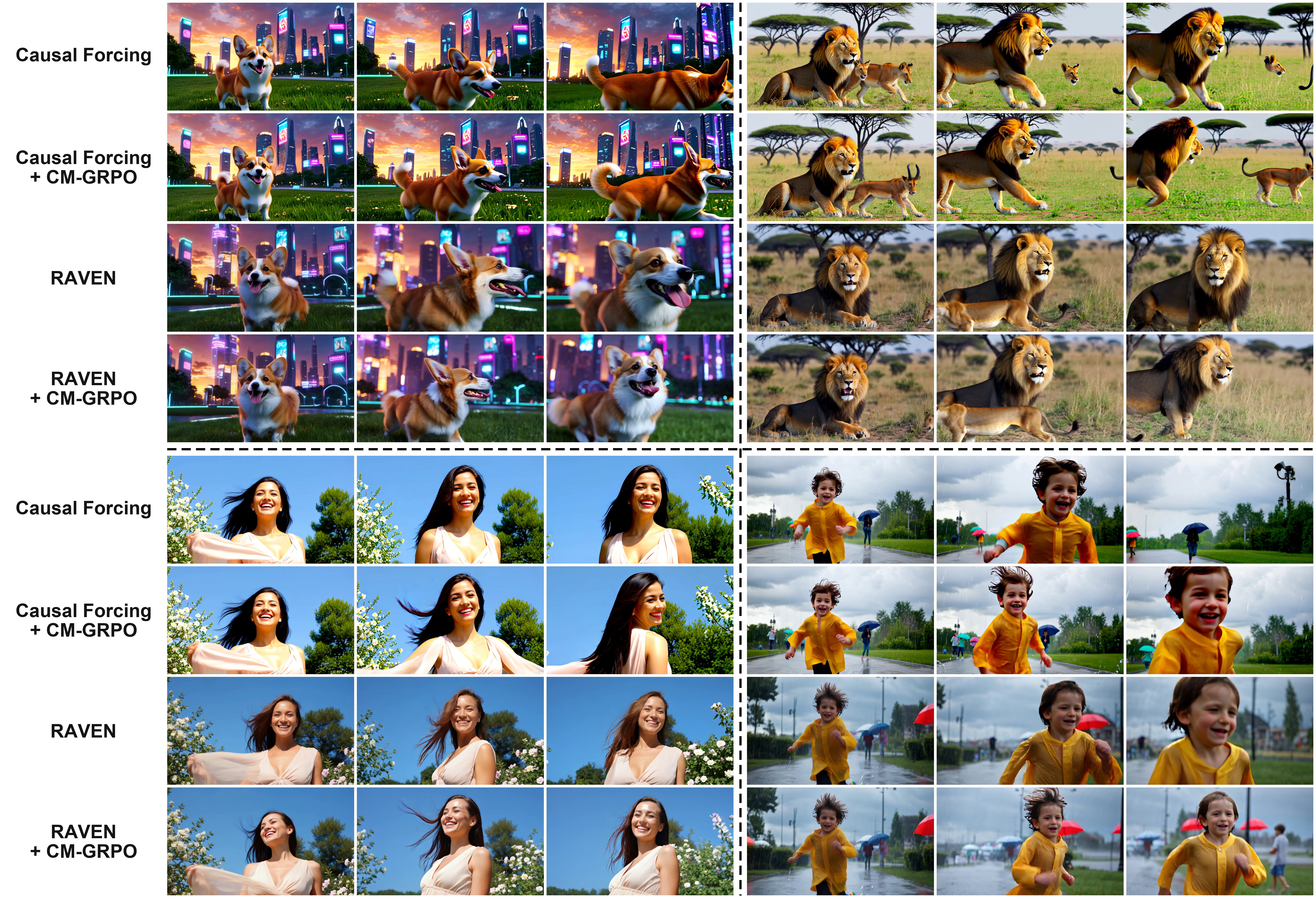}
    \vspace{-15pt}
    \caption{Qualitative comparisons. See supplementary for playble video clips.}
    \vspace{-10pt}
    \label{fig:qualitative_comparison}
\end{figure*}

\leavevmode
\begin{wraptable}[9]{r}{0.55\textwidth}
\vspace{-60pt}
\centering
\caption{Quantitative comparison results.}
\label{tab:quantitative_results}
\vspace{3pt}
\small
\setlength{\tabcolsep}{2pt}
\newcolumntype{M}{>{\raggedright\arraybackslash}X}
\newcolumntype{D}{>{\centering\arraybackslash}p{0.11\linewidth}}
\begin{tabularx}{\linewidth}{M D D D D}
\toprule[1.5pt]
\textbf{Method} &
\makecell[c]{\textbf{Total} \\ \textbf{Score}} &
\makecell[c]{\textbf{Qual.} \\ \textbf{Score}} &
\makecell[c]{\textbf{Sem.} \\ \textbf{Score}} &
\makecell[c]{\textbf{Dyn.} \\ \textbf{Deg.}} \\
\midrule[1pt]
CausVid~\cite{yin2025slowa} & 83.01 & 84.18 & 78.34 & 2.340 \\
LongLive~\cite{yang2025longlivea} & 83.05 & 83.70 & 80.46 & 2.277 \\
Rolling Forcing~\cite{liu2025rolling} & 83.25 & 84.00 & 80.25 & 2.536 \\
Self Forcing~\cite{huang2025selfa} & 84.27 & 85.10 & 80.97 & 2.543 \\
Reward Forcing~\cite{lu2025reward} & 84.39 & 85.27 & 80.87 & 2.508 \\
Causal Forcing~\cite{zhu2026causal} & 84.96 & 86.00 & 80.76 & 2.669 \\
\quad + CM-GRPO & 85.08 & 86.12 & 80.96 & 2.829 \\
\textbf{RAVEN} & 85.15 & 86.18 & 81.04 & 2.951 \\
\quad \textbf{+ CM-GRPO} & \textbf{85.46} & \textbf{86.54} & \textbf{81.17} & \textbf{2.962} \\
\bottomrule[1.5pt]
\end{tabularx}
\end{wraptable}
\noindent\textbf{Quantitative Comparisons.}
As reported in Table~\ref{tab:quantitative_results}, earlier causal distillation baselines cluster within a narrow range on the total score, and gains on quality or semantic alignment are typically offset by reduced motion or the reverse.
Causal Forcing~\cite{zhu2026causal} pushes the overall scores higher and partially recovers motion among the baselines, yet the same trade-off persists.
RAVEN surpasses every prior baseline across all four dimensions, with the largest margin on dynamic degree, indicating that supervising the cached history alleviates this trade-off rather than redistributing it.
Adding CM-GRPO to Causal Forcing yields a smaller gain concentrated on motion, whereas its combination with RAVEN produces the leading entry on every dimension, suggesting that the two contributions are complementary and that the policy update benefits from a generator already aligned with inference.

\noindent\textbf{Qualitative Comparisons.}
Figure~\ref{fig:qualitative_comparison} contrasts Causal Forcing and RAVEN, each with and without CM-GRPO, across four prompts spanning animal motion, urban scenery, and human subjects.
Causal Forcing exhibits severe structural failures under motion, stretching the corgi's body into an unnaturally long shape, detaching the lion's head from its body, and rendering colors in an over-saturated tone that looks unnatural.
Adding CM-GRPO repairs these structural breakdowns and the facial distortion of the running child, while RAVEN avoids the same failures from the start and produces more realistic colors and proportions, with only mild blurring around fast-moving regions such as the boy's face and the lion's body.
Combining RAVEN with CM-GRPO yields the most coherent results, sustaining structural stability and temporal smoothness through continuous motion such as the woman's turning hair and the child running in the rain.

\subsection{Ablation Studies}

\begin{figure*}[t]
    \centering
    \includegraphics[width=\linewidth]{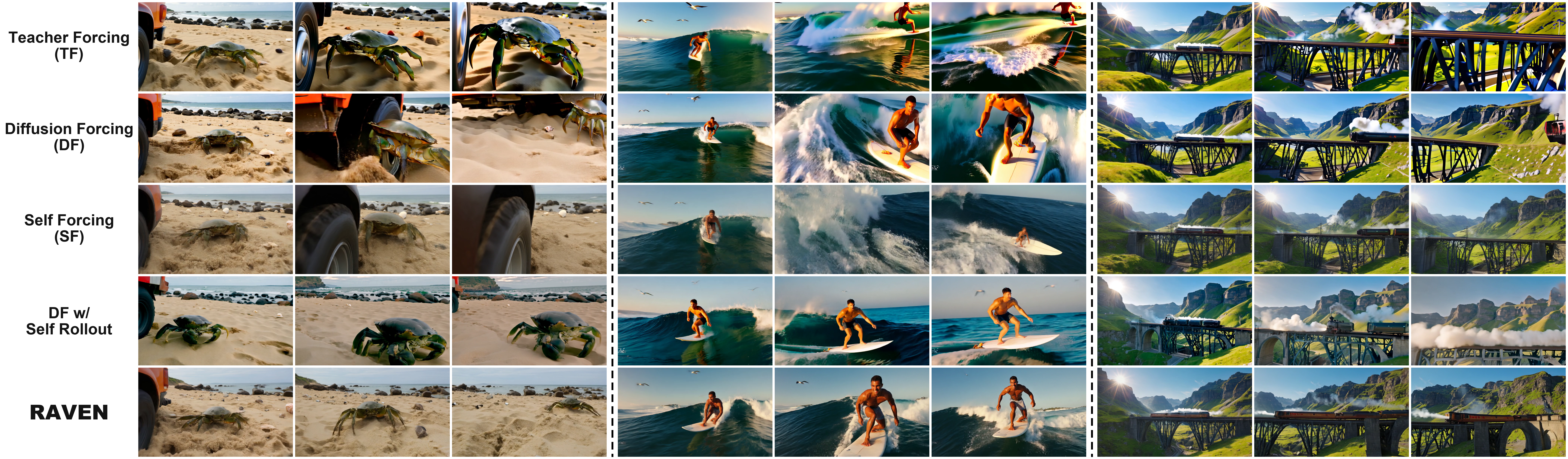}
    \vspace{-15pt}
    \caption{Qualitative ablation on Training-time Test. See supplementary for playable video clips.}
    \vspace{-10pt}
    \label{fig:ablation_qualitative}
\end{figure*}

\leavevmode
\begin{wraptable}[6]{r}{0.55\textwidth}
\vspace{-60pt}
\centering
\caption{Ablation on Training-time Test.}
\label{tab:training_time_test}
\vspace{3pt}
\small
\setlength{\tabcolsep}{2pt}
\begin{tabularx}{\linewidth}{>{\raggedright\arraybackslash}X *{4}{>{\centering\arraybackslash}p{0.11\linewidth}}}
\toprule[1.5pt]
\textbf{Method} &
\makecell[c]{\textbf{Total} \\ \textbf{Score}} &
\makecell[c]{\textbf{Qual.} \\ \textbf{Score}} &
\makecell[c]{\textbf{Sem.} \\ \textbf{Score}} &
\makecell[c]{\textbf{Dyn.} \\ \textbf{Deg.}} \\
\midrule[1pt]
Teacher Forcing (TF) & 82.64 & 83.11 & 80.73 & \textbf{3.000} \\
Diffusion Forcing (DF) & 84.09 & 84.75 & 81.45 & 2.743 \\
Self Forcing (SF) & 84.06 & 84.68 & \textbf{81.56} & 2.347 \\
DF w/ Self Rollout & 83.30 & 83.96 & 80.65 & 2.979 \\
\textbf{RAVEN} & \textbf{85.15} & \textbf{86.18} & 81.04 & 2.951 \\
\bottomrule[1.5pt]
\end{tabularx}
\vspace{5pt}
\end{wraptable}
\noindent\textbf{Effect of Training-time Test.}
All entries in Table~\ref{tab:training_time_test} share the same ODE-distilled initialization and chunk-wise loss scaling, isolating the effect of how history is formed and supervised.
TF achieves the strongest motion at the cost of the other two dimensions, while SF leads on semantic alignment but yields the weakest motion, reflecting how a detached cache withholds gradient from the history.
Replacing real prefixes in DF with self-rollout endpoints recovers motion close to TF yet erodes quality and semantic alignment, indicating that aligning the history distribution without supervising it merely redistributes the error.
RAVEN routes gradients from later chunks through the clean rollout endpoints used as history, attaining the leading total score while retaining motion close to TF.
These quantitative tradeoffs are visually validated in Figure~\ref{fig:ablation_qualitative}, where RAVEN consistently maintains structural coherence and object identity across frames, avoiding the severe structural distortions of TF and the blurring artifacts of SF.

\begin{figure*}[t]
    \centering
    \begin{minipage}[t]{0.44\textwidth}
        \vspace{-10pt}
        \centering
        \includegraphics[width=\linewidth]{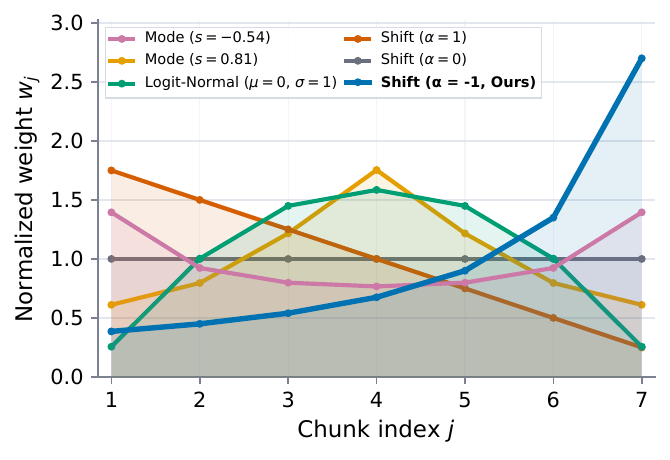}
    \end{minipage}
    \hfill
    \begin{minipage}[t]{0.55\textwidth}
        \vspace{0pt}
        \centering
        {\small
        \setlength{\tabcolsep}{2pt}
        \begin{tabularx}{\linewidth}{@{}>{\raggedright\arraybackslash}X *{4}{>{\centering\arraybackslash}p{0.11\linewidth}}@{}}
        \toprule[1.2pt]
        \textbf{Weighting Function $g_{\eta}$} &
        \makecell[c]{\textbf{Total} \\ \textbf{Score}} &
        \makecell[c]{\textbf{Qual.} \\ \textbf{Score}} &
        \makecell[c]{\textbf{Sem.} \\ \textbf{Score}} &
        \makecell[c]{\textbf{Dyn.} \\ \textbf{Deg.}} \\
        \midrule[0.8pt]
        Mode ($s=-0.54$) & 82.58 & 82.86 & \textbf{81.43} & 2.996 \\
        Mode ($s=0.81$) & 83.08 & 83.74 & 80.47 & 2.971 \\
        Logit-Normal ($\mu=0$, $\sigma=1$) & 83.31 & 83.97 & 80.65 & 2.963 \\
        \midrule[0.5pt]
        Shift ($\alpha=1$) & 83.79 & 84.87 & 79.46 & \textbf{3.000} \\
        Shift ($\alpha=0$) & 83.82 & 84.67 & 80.42 & 2.924 \\
        \textbf{Shift ($\boldsymbol{\alpha=-1}$, Ours)} & \textbf{85.15} & \textbf{86.18} & 81.04 & 2.951 \\
        \bottomrule[1.2pt]
        \end{tabularx}
        }
    \end{minipage}
    \vspace{-10pt}
    \caption{Ablation on Chunk-wise Loss Scaling.}
    \vspace{-10pt}
    \label{fig:chunk_wise_loss_scaling}
\end{figure*}

\noindent\textbf{Effect of Chunk-wise Loss Scaling.}
We compare candidate weighting functions $g_{\eta}$ that map the future participation score $p_j$ to a per-chunk weight, summarized in Figure~\ref{fig:chunk_wise_loss_scaling}.
Two candidates are inherited from SD3~\cite{esser2024scaling}, the mode density at $s \in \{-0.54, 0.81\}$ and the logit-normal density at $(\mu, \sigma) = (0, 1)$, each integrated over the participation interval covered by chunk $j$ to produce its raw weight.
The remaining family follows the shift parameterization $\pi_{\alpha}(p) = \alpha p / (1 + (\alpha - 1) p)$ borrowed from flow-matching timestep schedules, where $\alpha = 1$ biases mass toward early chunks and $\alpha = 0$ recovers a uniform schedule, while a negative $\alpha$ applies $\pi_{|\alpha|}$ to the reversed coordinate $p_{J}/p_{j}$ so that emphasis concentrates on later chunks instead.
Profiles peaked near the middle of the rollout or biased toward early chunks fall below the uniform baseline on total and quality, with the shift at $\alpha = 1$ trading semantic alignment for the highest dynamic degree.
The shift at $\alpha = -1$, the setting we adopt throughout RAVEN, instead raises the total score by roughly $1.3$ points while keeping motion and semantic alignment competitive.

% \leavevmode
\begin{wraptable}[12]{r}{0.55\textwidth}
\vspace{-5pt}
\centering
\caption{Ablation on Reward Composition.}
\label{tab:reward_composition}
\vspace{3pt}
\small
\setlength{\tabcolsep}{2pt}
\begin{tabularx}{\linewidth}{*{5}{>{\centering\arraybackslash}X} *{4}{>{\centering\arraybackslash}p{0.11\linewidth}}}
\toprule[1.5pt]
\multicolumn{5}{c}{\textbf{Reward Dimensions}} &
\multirow{2}[2]{*}{\makecell[c]{\textbf{Total} \\ \textbf{Score}}} &
\multirow{2}[2]{*}{\makecell[c]{\textbf{Qual.} \\ \textbf{Score}}} &
\multirow{2}[2]{*}{\makecell[c]{\textbf{Sem.} \\ \textbf{Score}}} &
\multirow{2}[2]{*}{\makecell[c]{\textbf{Dyn.} \\ \textbf{Deg.}}} \\
\cmidrule(lr){1-5}
\textbf{TA} & \textbf{DD} & \textbf{MS} & \textbf{AQ} & \textbf{IQ} & & & & \\
\midrule[1pt]
\multicolumn{5}{c}{RAVEN}          & 85.15 & 86.18 & 81.04 & 2.951 \\
\midrule[0.5pt]
1 & 0.35 & 0.75 & 1 & 1             & 84.82 & 85.77 & 80.99 & 2.913 \\
2 & 0.30 & 0.75 & 1 & 1             & 85.07 & 86.14 & 80.81 & 2.957 \\
2 & 0.35 & 1.00 & 1 & 1             & 85.24 & 86.27 & 81.14 & 2.936 \\
2 & 0.35 & 0.75 & 2 & 2             & 84.92 & 85.82 & \textbf{81.33} & 2.914 \\
\textbf{2} & \textbf{0.35} & \textbf{0.75} & \textbf{1} & \textbf{1} & \textbf{85.46} & \textbf{86.54} & 81.17 & \textbf{2.962} \\
\bottomrule[1.5pt]
\end{tabularx}
\vspace{5pt}
\end{wraptable}
\noindent\textbf{Effect of Reward Composition.}
Table~\ref{tab:reward_composition} ablates each reward dimension around our adopted composition.
Halving the \textit{Text Alignment} (TA) weight lowers the total and quality scores without a semantic gain, and reducing the \textit{Dynamic Degree} (DD) weight weakens both the total score and motion, indicating that the optical flow term carries the bulk of motion supervision rather than acting as a redundant complement to \textit{Motion Smoothness} (MS).
Raising the MS weight produces a smaller motion regression, while doubling \textit{Aesthetic Quality} (AQ) and \textit{Imaging Quality} (IQ) lifts semantic alignment to the highest entry at the cost of motion.
The adopted composition therefore best balances the three aspects identified in the methodology.

% \leavevmode
\begin{wraptable}[13]{r}{0.55\textwidth}
\vspace{-15pt}
\centering
\caption{Ablation on Consistency Policy.}
\label{tab:consistency_policy}
\vspace{3pt}
\small
\setlength{\tabcolsep}{2pt}
\begin{tabularx}{\linewidth}{>{\raggedright\arraybackslash}X *{4}{>{\centering\arraybackslash}p{0.11\linewidth}}}
\toprule[1.5pt]
\textbf{Policy Interface} &
\makecell[c]{\textbf{Total} \\ \textbf{Score}} &
\makecell[c]{\textbf{Qual.} \\ \textbf{Score}} &
\makecell[c]{\textbf{Sem.} \\ \textbf{Score}} &
\makecell[c]{\textbf{Dyn.} \\ \textbf{Deg.}} \\
\midrule[1pt]
RAVEN                                & 85.15 & 86.18 & 81.04 & 2.951 \\
+ EM ($\sigma=0.1$, $\beta=0$)        & 85.06 & 86.10 & 80.92 & 2.949 \\
+ EM ($\sigma=0.4$, $\beta=0$)        & 85.15 & 86.20 & 80.95 & 2.949 \\
+ EM ($\sigma=0.8$, $\beta=0$)        & 85.22 & 86.31 & 80.84 & 2.951 \\
+ EM ($\sigma=0.1$, $\beta=0.004$)    & 85.03 & 86.07 & 80.86 & 2.947 \\
+ EM ($\sigma=0.4$, $\beta=0.004$)    & 85.14 & 86.16 & 81.09 & 2.950 \\
+ EM ($\sigma=0.8$, $\beta=0.004$)    & 85.27 & 86.26 & \textbf{81.29} & 2.950 \\
\textbf{+ CM-GRPO (Ours)}             & \textbf{85.46} & \textbf{86.54} & 81.17 & \textbf{2.962} \\
\bottomrule[1.5pt]
\end{tabularx}
\vspace{5pt}
\end{wraptable}
\noindent\textbf{Effect of Consistency Policy.}
All entries in Table~\ref{tab:consistency_policy} share the RAVEN initialization and reward composition, differing only in the policy objective.
The Euler-Maruyama (EM) interfaces follow the discretization of Flow-GRPO~\cite{liu2025flowgrpo}, with $\sigma$ denoting the auxiliary diffusion noise level and $\beta$ the reference KL weight, and across the resulting grid the EM scores stay within a narrow range, with the strongest setting only marginally exceeding RAVEN.
CM-GRPO instead defines the policy objective on the consistency transition kernel used at inference and lifts the total, quality, and dynamic dimensions above every EM variant, surrendering only a small margin on semantic alignment.
This indicates that an auxiliary stochastic process is unnecessary once the policy kernel matches the inference sampler, which also relieves the joint tuning of the diffusion noise and KL weight.

\section{Conclusion}
We presented RAVEN, a training-time test framework that repacks each self rollout into an interleaved sequence of clean historical endpoints and noisy denoising states so that supervision propagates through the cached history used during autoregressive extrapolation.
We further proposed CM-GRPO, which formulates the policy objective directly on the consistency transition kernel used at inference rather than on an auxiliary Euler-Maruyama process.
Together they surpass recent causal video distillation baselines across quality, semantic, and dynamic dimensions, with their combination yielding the strongest results on every metric considered.
Both formulations also admit broader scope than the setting evaluated here, and we discuss their generalizability in greater detail in Appendix~\ref{app:discussion}.

\section{Acknowledgements}
We thank Fei Ni and Changrui Chen for their valuable discussions and assistance.
J. Deng was supported by the NVIDIA Academic Grant.
The authors acknowledge the use of resources provided by the Isambard-AI National AI Research Resource (AIRR). Isambard-AI~\cite{mcintosh2024isambard} is operated by the University of Bristol and is funded by the UK Government’s Department for Science, Innovation and Technology (DSIT) via UK Research and Innovation; and the Science and Technology Facilities Council [ST/AIRR/I-A-I/1023].

{
    % \newpage
    \small
    \bibliographystyle{unsrtnat}
    \bibliography{main}

@article{ai2025magi1a,
  title = {MAGI-1: Autoregressive Video Generation at Scale},
  year = 2025,
  journal = {arXiv preprint arXiv:2505.13211},
  author = {ai, Sand and Teng, Hansi and Jia, Hongyu and Sun, Lei and Li, Lingzhi and Li, Maolin and Tang, Mingqiu and Han, Shuai and Zhang, Tianning and Zhang, W. Q. and Luo, Weifeng and Kang, Xiaoyang and others}
}

@article{arkhipkin2025kandinsky,
  title = {Kandinsky 5.0: A Family of Foundation Models for Image and Video Generation},
  year = 2025,
  journal = {arXiv preprint arXiv:2511.14993},
  author = {Arkhipkin, Vladimir and Korviakov, Vladimir and Gerasimenko, Nikolai and Parkhomenko, Denis and Vasilev, Viacheslav and Letunovskiy, Alexey and Vaulin, Nikolai and Kovaleva, Maria and Kirillov, Ivan and Novitskiy, Lev and Koposov, Denis and Kiselev, Nikita and others}
}

@article{bai2026causality,
  title = {Causality in Video Diffusers is Separable from Denoising},
  author = {Bai, Xingjian and He, Guande and Li, Zhengqi and Shechtman, Eli and Huang, Xun and Wu, Zongze},
  year = 2026,
  journal = {arXiv preprint arXiv:2602.10095}
}

@inproceedings{black2024training,
  title = {Training Diffusion Models with Reinforcement Learning},
  booktitle = {ICLR},
  author = {Black, Kevin and Janner, Michael and Du, Yilun and Kostrikov, Ilya and Levine, Sergey},
  year = 2024
}

@inproceedings{chen2024diffusiona,
  title = {Diffusion Forcing: Next-token Prediction Meets Full-Sequence Diffusion},
  booktitle = {NeurIPS},
  author = {Chen, Boyuan and Monso, Diego Marti and Du, Yilun and Simchowitz, Max and Tedrake, Russ and Sitzmann, Vincent},
  year = 2024
}

@article{chen2025flashdmd,
  title = {Flash-DMD: Towards High-Fidelity Few-Step Image Generation with Efficient Distillation and Joint Reinforcement Learning},
  author = {Chen, Guanjie and Huang, Shirui and Liu, Kai and Zhu, Jianchen and Qu, Xiaoye and Chen, Peng and Cheng, Yu and Sun, Yifu},
  year = 2025,
  journal = {arXiv preprint arXiv:2511.20549}
}

@article{chen2025skyreelsv2a,
  title = {SkyReels-V2: Infinite-length Film Generative Model},
  year = 2025,
  journal = {arXiv preprint arXiv:2504.13074},
  author = {Chen, Guibin and Lin, Dixuan and Yang, Jiangping and Lin, Chunze and Zhu, Junchen and Fan, Mingyuan and Zhang, Hao and Chen, Sheng and Chen, Zheng and Ma, Chengcheng and Xiong, Weiming and Wang, Wei and others}
}

@article{chen2026context,
  title = {Context Forcing: Consistent Autoregressive Video Generation with Long Context},
  author = {Chen, Shuo and Wei, Cong and Sun, Sun and Nie, Ping and Zhou, Kai and Zhang, Ge and Yang, Ming-Hsuan and Chen, Wenhu},
  year = 2026,
  journal = {arXiv preprint arXiv:2602.06028}
}

@article{chen2026futureinformed,
  title = {Past- and Future-Informed KV Cache Policy with Salience Estimation in Autoregressive Video Diffusion},
  author = {Chen, Hanmo and Xu, Chenghao and Yang, Xu and Chen, Xuan and Deng, Cheng},
  year = 2026,
  journal = {arXiv preprint arXiv:2601.21896}
}

@inproceedings{chen2026sanavideo,
  title = {SANA-Video: Efficient Video Generation with Block Linear Diffusion Transformer},
  booktitle = {ICLR},
  year = 2026,
  author = {Chen, Junsong and Zhao, Yuyang and Yu, Jincheng and Chu, Ruihang and Chen, Junyu and Yang, Shuai and Wang, Xianbang and Pan, Yicheng and Zhou, Daquan and Ling, Huan and Liu, Haozhe and Yi, Hongwei and others}
}

@article{chen2026superflow,
  title = {SuperFlow: Training Flow Matching Models with RL on the Fly},
  author = {Chen, Kaijie and Xu, Zhiyang and Shen, Ying and Lin, Zihao and Yao, Yuguang and Huang, Lifu},
  year = 2026,
  journal = {arXiv preprint arXiv:2512.17951}
}

@article{cui2025selfforcinga,
  title = {Self-Forcing++: Towards Minute-Scale High-Quality Video Generation},
  author = {Cui, Justin and Wu, Jie and Li, Ming and Yang, Tao and Li, Xiaojie and Wang, Rui and Bai, Andrew and Ban, Yuanhao and Hsieh, Cho-Jui},
  year = 2025,
  journal = {arXiv preprint arXiv:2510.02283}
}

@article{cui2026lol,
  title = {LoL: Longer than Longer, Scaling Video Generation to Hour},
  author = {Cui, Justin and Wu, Jie and Li, Ming and Yang, Tao and Li, Xiaojie and Wang, Rui and Bai, Andrew and Ban, Yuanhao and Hsieh, Cho-Jui},
  year = 2026,
  journal = {arXiv preprint arXiv:2601.16914}
}

@inproceedings{deng2025autoregressive,
  title = {Autoregressive Video Generation without Vector Quantization},
  booktitle = {ICLR},
  author = {Deng, Haoge and Pan, Ting and Diao, Haiwen and Luo, Zhengxiong and Cui, Yufeng and Lu, Huchuan and Shan, Shiguang and Qi, Yonggang and Wang, Xinlong},
  year = 2025
}

@inproceedings{ding2025treegrpo,
  title = {TreeGRPO: Tree-Advantage GRPO for Online RL Post-Training of Diffusion Models},
  booktitle = {ICLR},
  author = {Ding, Zheng and Ye, Weirui},
  year = 2026
}

@article{fu2025dynamictreerpo,
  title = {Dynamic-TreeRPO: Breaking the Independent Trajectory Bottleneck with Structured Sampling},
  year = 2025,
  journal = {arXiv preprint arXiv:2509.23352},
  author = {Fu, Xiaolong and Ma, Lichen and Guo, Zipeng and Zhou, Gaojing and Wang, Chongxiao and Dong, ShiPing and Zhou, Shizhe and Zhou, Shizhe and Liu, Ximan and Fu, Jingling and Sin, Tan Lit and Shi, Yu and others}
}

@article{gao2025seedance,
  title = {Seedance 1.0: Exploring the Boundaries of Video Generation Models},
  year = 2025,
  journal = {arXiv preprint arXiv:2506.09113},
  author = {Gao, Yu and Guo, Haoyuan and Hoang, Tuyen and Huang, Weilin and Jiang, Lu and Kong, Fangyuan and Li, Huixia and Li, Jiashi and Li, Liang and Li, Xiaojie and Li, Xunsong and Li, Yifu and others}
}

@misc{genmo2024mochi,
  title = {Mochi 1: AI Video Generator},
  author = {{Genmo}},
  year = 2024,
  howpublished = {\url{https://www.genmo.ai/blog/mochi-1-a-new-sota-in-open-text-to-video}}
}

@article{guo2025endtoend,
  title = {End-to-End Training for Autoregressive Video Diffusion via Self-Resampling},
  author = {Guo, Yuwei and Yang, Ceyuan and He, Hao and Zhao, Yang and Wei, Meng and Yang, Zhenheng and Huang, Weilin and Lin, Dahua},
  year = 2025,
  journal = {arXiv preprint arXiv:2512.15702}
}

@article{guo2026erudiff,
  title = {EruDiff: Refactoring Knowledge in Diffusion Models for Advanced Text-to-Image Synthesis},
  author = {Guo, Xiefan and Ma, Xinzhu and Ma, Haoxiang and Zhou, Zihao and Huang, Di},
  year = 2026,
  journal = {arXiv preprint arXiv:2603.20828}
}

@article{hacohen2024ltxvideob,
  title = {LTX-Video: Realtime Video Latent Diffusion},
  year = 2024,
  journal = {arXiv preprint arXiv:2501.00103},
  author = {HaCohen, Yoav and Chiprut, Nisan and Brazowski, Benny and Shalem, Daniel and Moshe, Dudu and Richardson, Eitan and Levin, Eran and Shiran, Guy and Zabari, Nir and Gordon, Ori and Panet, Poriya and Weissbuch, Sapir and others}
}

@article{hacohen2026ltx2,
  title = {LTX-2: Efficient Joint Audio-Visual Foundation Model},
  year = 2026,
  journal = {arXiv preprint arXiv:2601.03233},
  author = {HaCohen, Yoav and Brazowski, Benny and Chiprut, Nisan and Bitterman, Yaki and Kvochko, Andrew and Berkowitz, Avishai and Shalem, Daniel and Lifschitz, Daphna and Moshe, Dudu and Porat, Eitan and Richardson, Eitan and Shiran, Guy and others}
}

@article{he2025gardo,
  title = {GARDO: Reinforcing Diffusion Models without Reward Hacking},
  author = {He, Haoran and Ye, Yuxiao and Liu, Jie and Liang, Jiajun and Wang, Zhiyong and Yuan, Ziyang and Wang, Xintao and Mao, Hangyu and Wan, Pengfei and Pan, Ling},
  year = 2025,
  journal = {arXiv preprint arXiv:2512.24138}
}

@inproceedings{he2025tempflowgrpo,
  title = {TempFlow-GRPO: When Timing Matters for GRPO in Flow Models},
  booktitle = {ICLR},
  author = {He, Xiaoxuan and Fu, Siming and Zhao, Yuke and Li, Wanli and Yang, Jian and Yin, Dacheng and Rao, Fengyun and Zhang, Bo},
  year = 2026
}

@inproceedings{he2026neighbor,
  title = {Neighbor GRPO: Contrastive ODE Policy Optimization Aligns Flow Models},
  booktitle = {CVPR},
  author = {He, Dailan and Feng, Guanlin and Ge, Xingtong and Niu, Yazhe and Zhang, Yi and Ma, Bingqi and Song, Guanglu and Liu, Yu and Li, Hongsheng},
  year = 2026
}

@inproceedings{huang2022realtime,
  title = {Real-Time Intermediate Flow Estimation for Video Frame Interpolation},
  booktitle = {ECCV},
  author = {Huang, Zhewei and Zhang, Tianyuan and Heng, Wen and Shi, Boxin and Zhou, Shuchang},
  year = 2022
}

@inproceedings{huang2024vbenchb,
  title = {VBench: Comprehensive Benchmark Suite for Video Generative Models},
  booktitle = {CVPR},
  year = 2024,
  author = {Huang, Ziqi and He, Yinan and Yu, Jiashuo and Zhang, Fan and Si, Chenyang and Jiang, Yuming and Zhang, Yuanhan and Wu, Tianxing and Jin, Qingyang and Chanpaisit, Nattapol and Wang, Yaohui and Chen, Xinyuan and others}
}

@inproceedings{huang2025selfa,
  title = {Self Forcing: Bridging the Train-Test Gap in Autoregressive Video Diffusion},
  booktitle = {NeurIPS},
  author = {Huang, Xun and Li, Zhengqi and He, Guande and Zhou, Mingyuan and Shechtman, Eli},
  year = 2025
}

@article{huang2026live,
  title = {LIVE: Long-horizon Interactive Video World Modeling},
  author = {Huang, Junchao and Ye, Ziyang and Hu, Xinting and He, Tianyu and Zhang, Guiyu and Shi, Shaoshuai and Bian, Jiang and Jiang, Li},
  year = 2026,
  journal = {arXiv preprint arXiv:2602.03747}
}

@article{ji2025memflow,
  title = {MemFlow: Flowing Adaptive Memory for Consistent and Efficient Long Video Narratives},
  author = {Ji, Sihui and Chen, Xi and Yang, Shuai and Tao, Xin and Wan, Pengfei and Zhao, Hengshuang},
  year = 2025,
  journal = {arXiv preprint arXiv:2512.14699}
}

@inproceedings{jin2024pyramidal,
  title = {Pyramidal Flow Matching for Efficient Video Generative Modeling},
  booktitle = {ICLR},
  author = {Jin, Yang and Sun, Zhicheng and Li, Ningyuan and Xu, Kun and Xu, Kun and Jiang, Hao and Zhuang, Nan and Huang, Quzhe and Song, Yang and Mu, Yadong and Lin, Zhouchen},
  year = 2025
}

@inproceedings{ke2021musiq,
  title = {MUSIQ: Multi-scale Image Quality Transformer},
  booktitle = {ICCV},
  author = {Ke, Junjie and Wang, Qifei and Wang, Yilin and Milanfar, Peyman and Yang, Feng},
  year = 2021,
  pages = {5128--5137}
}

@inproceedings{kim2024consistency,
  title = {Consistency Trajectory Models: Learning Probability Flow ODE Trajectory of Diffusion},
  booktitle = {ICLR},
  author = {Kim, Dongjun and Lai, Chieh-Hsin and Liao, Wei-Hsiang and Murata, Naoki and Takida, Yuhta and Uesaka, Toshimitsu and He, Yutong and Mitsufuji, Yuki and Ermon, Stefano},
  year = 2024
}

@article{kong2025hunyuanvideo,
  title = {HunyuanVideo: A Systematic Framework For Large Video Generative Models},
  year = 2025,
  journal = {arXiv preprint arXiv:2412.03603},
  author = {Kong, Weijie and Tian, Qi and Zhang, Zijian and Min, Rox and Dai, Zuozhuo and Zhou, Jin and Xiong, Jiangfeng and Li, Xin and Wu, Bo and Zhang, Jianwei and Wu, Kathrina and Lin, Qin and others}
}

@misc{laion-ai2022aestheticpredictor,
  title = {aesthetic-predictor},
  author = {{LAION-AI}},
  year = 2022,
  howpublished = {\url{https://github.com/LAION-AI/aesthetic-predictor}}
}

@inproceedings{li2023amt,
  title = {AMT: All-Pairs Multi-Field Transforms for Efficient Frame Interpolation},
  booktitle = {CVPR},
  author = {Li, Zhen and Zhu, Zuo-Liang and Han, Ling-Hao and Hou, Qibin and Guo, Chun-Le and Cheng, Ming-Ming},
  year = 2023,
  pages = {9801--9810}
}

@inproceedings{li2025branchgrpo,
  title = {BranchGRPO: Stable and Efficient GRPO with Structured Branching in Diffusion Models},
  booktitle = {ICLR},
  author = {Li, Yuming and Wang, Yikai and Zhu, Yuying and Zhao, Zhongyu and Lu, Ming and She, Qi and Zhang, Shanghang},
  year = 2026
}

@inproceedings{li2025eagle3,
  title = {EAGLE-3: Scaling up Inference Acceleration of Large Language Models via Training-Time Test},
  booktitle = {NeurIPS},
  author = {Li, Yuhui and Wei, Fangyun and Zhang, Chao and Zhang, Hongyang},
  year = 2025
}

@inproceedings{li2025stable,
  title = {Stable Video Infinity: Infinite-Length Video Generation with Error Recycling},
  booktitle = {ICLR},
  author = {Li, Wuyang and Pan, Wentao and Luan, Po-Chien and Gao, Yang and Alahi, Alexandre},
  year = 2026
}

@article{li2026train,
  title = {Train Short, Inference Long: Training-free Horizon Extension for Autoregressive Video Generation},
  author = {Li, Jia and Fu, Xiaomeng and Peng, Xurui and Chen, Weifeng and Zheng, Youwei and Zhao, Tianyu and Wang, Jiexi and Chen, Fangmin and Wang, Xing and So, Hayden Kwok-Hay},
  year = 2026,
  journal = {arXiv preprint arXiv:2602.14027}
}

@article{liang2026leapalign,
  title = {LeapAlign: Post-Training Flow Matching Models at Any Generation Step by Building Two-Step Trajectories},
  author = {Liang, Zhanhao and Yang, Tao and Wu, Jie and Feng, Chengjian and Zheng, Liang},
  year = 2026,
  journal = {arXiv preprint arXiv:2604.15311}
}

@article{lin2024sdxllightning,
  title = {SDXL-Lightning: Progressive Adversarial Diffusion Distillation},
  author = {Lin, Shanchuan and Wang, Anran and Yang, Xiao},
  year = 2024,
  journal = {arXiv preprint arXiv:2402.13929}
}

@inproceedings{liu2025flowgrpo,
  title = {Flow-GRPO: Training Flow Matching Models via Online RL},
  booktitle = {NeurIPS},
  author = {Liu, Jie and Liu, Gongye and Liang, Jiajun and Li, Yangguang and Liu, Jiaheng and Wang, Xintao and Wan, Pengfei and Zhang, Di and Ouyang, Wanli},
  year = 2025
}

@inproceedings{liu2025improving,
  title = {Improving Video Generation with Human Feedback},
  booktitle = {NeurIPS},
  year = 2025,
  author = {Liu, Jie and Liu, Gongye and Liang, Jiajun and Yuan, Ziyang and Liu, Xiaokun and Zheng, Mingwu and Wu, Xiele and Wang, Qiulin and Xia, Menghan and Wang, Xintao and Liu, Xiaohong and Yang, Fei and others}
}

@inproceedings{liu2025infinitystar,
  title = {InfinityStar: Unified Spacetime AutoRegressive Modeling for Visual Generation},
  booktitle = {NeurIPS},
  author = {Liu, Jinlai and Han, Jian and Yan, Bin and Wu, Hui and Zhu, Fengda and Wang, Xing and Jiang, Yi and Peng, Bingyue and Yuan, Zehuan},
  year = 2025
}

@inproceedings{liu2025rolling,
  title = {Rolling Forcing: Autoregressive Long Video Diffusion in Real Time},
  booktitle = {ICLR},
  author = {Liu, Kunhao and Hu, Wenbo and Xu, Jiale and Shan, Ying and Lu, Shijian},
  year = 2026
}

@inproceedings{liu2026streaming,
  title = {Streaming Autoregressive Video Generation via Diagonal Distillation},
  booktitle = {ICLR},
  author = {Liu, Jinxiu and Liu, Xuanming and Mei, Kangfu and Wen, Yandong and Yang, Ming-Hsuan and Liu, Weiyang},
  year = 2026
}

@article{liu2026unigrpo,
  title = {UniGRPO: Unified Policy Optimization for Reasoning-Driven Visual Generation},
  author = {Liu, Jie and Ye, Zilyu and Yuan, Linxiao and Zhu, Shenhan and Gao, Yu and Wu, Jie and Li, Kunchang and Wang, Xionghui and Nie, Xiaonan and Huang, Weilin and Ouyang, Wanli},
  year = 2026,
  journal = {arXiv preprint arXiv:2603.23500}
}

@article{lu2025hyperbagel,
  title = {Hyper-Bagel: A Unified Acceleration Framework for Multimodal Understanding and Generation},
  author = {Lu, Yanzuo and Xia, Xin and Zhang, Manlin and Kuang, Huafeng and Zheng, Jianbin and Ren, Yuxi and Xiao, Xuefeng},
  year = 2025,
  journal = {arXiv preprint arXiv:2509.18824}
}

@inproceedings{lu2025reward,
  title = {Reward Forcing: Efficient Streaming Video Generation with Rewarded Distribution Matching Distillation},
  booktitle = {CVPR},
  author = {Lu, Yunhong and Zeng, Yanhong and Li, Haobo and Ouyang, Hao and Wang, Qiuyu and Cheng, Ka Leong and Zhu, Jiapeng and Cao, Hengyuan and Zhang, Zhipeng and Zhu, Xing and Shen, Yujun and Zhang, Min},
  year = 2026
}

@inproceedings{luadversarial,
  title = {Adversarial Distribution Matching for Diffusion Distillation Towards Efficient Image and Video Synthesis},
  booktitle = {ICCV},
  author = {Lu, Yanzuo and Ren, Yuxi and Xia, Xin and Lin, Shanchuan and Wang, Xing and Xiao, Xuefeng and Ma, Andy J and Xie, Xiaohua and Lai, Jian-Huang},
  year = 2025
}

@article{luo2023latent,
  title = {Latent Consistency Models: Synthesizing High-Resolution Images with Few-Step Inference},
  author = {Luo, Simian and Tan, Yiqin and Huang, Longbo and Li, Jian and Zhao, Hang},
  year = 2023,
  journal = {arXiv preprint arXiv:2310.04378}
}

@article{luo2023lcmlora,
  title = {LCM-LoRA: A Universal Stable-Diffusion Acceleration Module},
  author = {Luo, Simian and Tan, Yiqin and Patil, Suraj and Gu, Daniel and von Platen, Patrick and Passos, Apolin{\'a}rio and Huang, Longbo and Li, Jian and Zhao, Hang},
  year = 2023,
  journal = {arXiv preprint arXiv:2311.05556}
}

@inproceedings{luo2024onestep,
  title = {One-Step Diffusion Distillation through Score Implicit Matching},
  booktitle = {NeurIPS},
  author = {Luo, Weijian and Huang, Zemin and Geng, Zhengyang and Kolter, J. Zico and Qi, Guo-jun},
  year = 2024
}

@inproceedings{luo2025reinforcing,
  title = {Reinforcing Diffusion Models by Direct Group Preference Optimization},
  booktitle = {ICLR},
  author = {Luo, Yihong and Hu, Tianyang and Tang, Jing},
  year = 2026
}

@article{luo2025sample,
  title = {Sample By Step, Optimize By Chunk: Chunk-Level GRPO For Text-to-Image Generation},
  author = {Luo, Yifu and Du, Penghui and Li, Bo and Du, Sinan and Zhang, Tiantian and Chang, Yongzhe and Wu, Kai and Gai, Kun and Wang, Xueqian},
  year = 2025,
  journal = {arXiv preprint arXiv:2510.21583}
}

@article{luo2026tdmr1,
  title = {TDM-R1: Reinforcing Few-Step Diffusion Models with Non-Differentiable Reward},
  author = {Luo, Yihong and Hu, Tianyang and Luo, Weijian and Tang, Jing},
  year = 2026,
  journal = {arXiv preprint arXiv:2603.07700}
}

@article{lyu2025multigrpo,
  title = {Multi-GRPO: Multi-Group Advantage Estimation for Text-to-Image Generation with Tree-Based Trajectories and Multiple Rewards},
  author = {Lyu, Qiang and Chen, Zicong and Wang, Chongxiao and Shi, Haolin and Gao, Shibo and Piao, Ran and Zeng, Youwei and Si, Jianlou and Ding, Fei and Li, Jing and Lau, Chun Pong and Wang, Weiqiang},
  year = 2025,
  journal = {arXiv preprint arXiv:2512.00743}
}

@article{ma2025stage,
  title = {STAGE: Stable and Generalizable GRPO for Autoregressive Image Generation},
  author = {Ma, Xiaoxiao and Qiu, Haibo and Zhang, Guohui and Zeng, Zhixiong and Yang, Siqi and Ma, Lin and Zhao, Feng},
  year = 2025,
  journal = {arXiv preprint arXiv:2509.25027}
}

@article{ma2025stepvideot2v,
  title = {Step-Video-T2V Technical Report: The Practice, Challenges, and Future of Video Foundation Model},
  year = 2025,
  journal = {arXiv preprint arXiv:2502.10248},
  author = {Ma, Guoqing and Huang, Haoyang and Yan, Kun and Chen, Liangyu and Duan, Nan and Yin, Shengming and Wan, Changyi and Ming, Ranchen and Song, Xiaoniu and Chen, Xing and Zhou, Yu and Sun, Deshan and others}
}

@article{mcallister2026finite,
  title = {Finite Difference Flow Optimization for RL Post-Training of Text-to-Image Models},
  author = {McAllister, David and Aittala, Miika and Karras, Tero and Hellsten, Janne and Kanazawa, Angjoo and Aila, Timo and Laine, Samuli},
  year = 2026,
  journal = {arXiv preprint arXiv:2603.12893}
}

@misc{meta2024movie,
  title = {Movie Gen: A Cast of Media Foundation Models},
  author = {{Meta}},
  year = 2024,
  howpublished = {\url{https://ai.meta.com/static-resource/movie-gen-research-paper}}
}

@inproceedings{nan2025openvid1ma,
  title = {OpenVid-1M: A Large-Scale High-Quality Dataset for Text-to-video Generation},
  booktitle = {ICLR},
  author = {Nan, Kepan and Xie, Rui and Zhou, Penghao and Fan, Tiehan and Yang, Zhenheng and Chen, Zhijie and Li, Xiang and Yang, Jian and Tai, Ying},
  year = 2025
}

@misc{nvidia2025cosmos,
  title = {Cosmos World Foundation Model Platform for Physical AI},
  author = {{NVIDIA}},
  year = 2025,
  howpublished = {\url{https://research.nvidia.com/publication/2025-01_cosmos-world-foundation-model-platform-physical-ai}}
}

@article{po2025bagger,
  title = {BAgger: Backwards Aggregation for Mitigating Drift in Autoregressive Video Diffusion Models},
  author = {Po, Ryan and Chan, Eric Ryan and Chen, Changan and Wetzstein, Gordon},
  year = 2025,
  journal = {arXiv preprint arXiv:2512.12080}
}

@inproceedings{rafailov2024direct,
  title = {Direct Preference Optimization: Your Language Model is Secretly a Reward Model},
  booktitle = {NeurIPS},
  author = {Rafailov, Rafael and Sharma, Archit and Mitchell, Eric and Ermon, Stefano and Manning, Christopher D. and Finn, Chelsea},
  year = 2023
}

@inproceedings{ren2024hypersd,
  title = {Hyper-SD: Trajectory Segmented Consistency Model for Efficient Image Synthesis},
  booktitle = {NeurIPS},
  author = {Ren, Yuxi and Xia, Xin and Lu, Yanzuo and Zhang, Jiacheng and Wu, Jie and Xie, Pan and Wang, Xing and Xiao, Xuefeng},
  year = 2024
}

@inproceedings{salimans2022progressive,
  title = {Progressive Distillation for Fast Sampling of Diffusion Models},
  booktitle = {ICLR},
  author = {Salimans, Tim and Ho, Jonathan},
  year = 2022
}

@inproceedings{savani2026stepwise,
  title = {Stepwise Credit Assignment for GRPO on Flow-Matching Models},
  booktitle = {CVPR},
  author = {Savani, Yash and Kveton, Branislav and Liu, Yuchen and Wang, Yilin and Shi, Jing and Mukherjee, Subhojyoti and Vlassis, Nikos and Singh, Krishna Kumar},
  year = 2026
}

@article{seedance2025seedance,
  title = {Seedance 1.5 pro: A Native Audio-Visual Joint Generation Foundation Model},
  year = 2025,
  journal = {arXiv preprint arXiv:2512.13507},
  author = {Seedance, Team and Chen, Heyi and Chen, Siyan and Chen, Xin and Chen, Yanfei and Chen, Ying and Chen, Zhuo and Cheng, Feng and Cheng, Tianheng and Cheng, Xinqi and Chi, Xuyan and Cong, Jian and others}
}

@article{seedance2026seedance,
  title = {Seedance 2.0: Advancing Video Generation for World Complexity},
  year = 2026,
  journal = {arXiv preprint arXiv:2604.14148},
  author = {Seedance, Team and Chen, De and Chen, Liyang and Chen, Xin and Chen, Ying and Chen, Zhuo and Chen, Zhuowei and Cheng, Feng and Cheng, Tianheng and Cheng, Yufeng and Chi, Mojie and Chi, Xuyan and others}
}

@article{shao2024deepseekmath,
  title = {DeepSeekMath: Pushing the Limits of Mathematical Reasoning in Open Language Models},
  author = {Shao, Zhihong and Wang, Peiyi and Zhu, Qihao and Xu, Runxin and Song, Junxiao and Bi, Xiao and Zhang, Haowei and Zhang, Mingchuan and Li, Y. K. and Wu, Y. and Guo, Daya},
  year = 2024,
  journal = {arXiv preprint arXiv:2402.03300}
}

@article{shao2025anchoring,
  title = {Anchoring Values in Temporal and Group Dimensions for Flow Matching Model Alignment},
  author = {Shao, Yawen and Xiao, Jie and Zhu, Kai and Liu, Yu and Zhai, Wei and Cao, Yang and Zha, Zheng-Jun},
  year = 2025,
  journal = {arXiv preprint arXiv:2512.12387}
}

@article{sheng2025understanding,
  title = {Understanding Sampler Stochasticity in Training Diffusion Models for RLHF},
  author = {Sheng, Jiayuan and Zhao, Hanyang and Chen, Haoxian and Yao, David D. and Tang, Wenpin},
  year = 2025,
  journal = {arXiv preprint arXiv:2510.10767}
}

@inproceedings{shin2026motionstream,
  title = {MotionStream: Real-Time Video Generation with Interactive Motion Controls},
  booktitle = {ICLR},
  author = {Shin, Joonghyuk and Li, Zhengqi and Zhang, Richard and Zhu, Jun-Yan and Park, Jaesik and Shechtman, Eli and Huang, Xun},
  year = 2026
}

@inproceedings{song2023consistency,
  title = {Consistency Models},
  booktitle = {ICML},
  author = {Song, Yang and Dhariwal, Prafulla and Chen, Mark and Sutskever, Ilya},
  year = 2023
}

@inproceedings{song2025historyguided,
  title = {History-Guided Video Diffusion},
  booktitle = {ICML},
  author = {Song, Kiwhan and Chen, Boyuan and Simchowitz, Max and Du, Yilun and Tedrake, Russ and Sitzmann, Vincent},
  year = 2025
}

@article{tang2025tr2d2,
  title = {TR2-D2: Tree Search Guided Trajectory-Aware Fine-Tuning for Discrete Diffusion},
  author = {Tang, Sophia and Zhu, Yuchen and Tao, Molei and Chatterjee, Pranam},
  year = 2025,
  journal = {arXiv preprint arXiv:2509.25171}
}

@inproceedings{teed2020raft,
  title = {RAFT: Recurrent All-Pairs Field Transforms for Optical Flow},
  booktitle = {ECCV},
  author = {Teed, Zachary and Deng, Jia},
  year = 2020,
  volume = {12347},
  pages = {402--419}
}

@article{tong2026alleviating,
  title = {Alleviating Sparse Rewards by Modeling Step-Wise and Long-Term Sampling Effects in Flow-Based GRPO},
  author = {Tong, Yunze and Liu, Mushui and Zhao, Canyu and He, Wanggui and Zhang, Shiyi and Zhang, Hongwei and Zhang, Peng and Liu, Jinlong and Huang, Ju and Wang, Jiamang and Jiang, Hao and Huang, Pipei},
  year = 2026,
  journal = {arXiv preprint arXiv:2602.06422}
}

@article{wan2025wan,
  title = {Wan: Open and Advanced Large-Scale Video Generative Models},
  year = 2025,
  journal = {arXiv preprint arXiv:2503.20314},
  author = {Wan, Team and Wang, Ang and Ai, Baole and Wen, Bin and Mao, Chaojie and Xie, Chen-Wei and Chen, Di and Yu, Feiwu and Zhao, Haiming and Yang, Jianxiao and Zeng, Jianyuan and Wang, Jiayu and others}
}

@inproceedings{wang2024phased,
  title = {Phased Consistency Model},
  booktitle = {NeurIPS},
  author = {Wang, Fu-Yun and Huang, Zhaoyang and Bergman, Alexander William and Shen, Dazhong and Gao, Peng and Lingelbach, Michael and Sun, Keqiang and Bian, Weikang and Song, Guanglu and Liu, Yu and Li, Hongsheng and Wang, Xiaogang},
  year = 2024
}

@article{wang2024qwen2vl,
  title = {Qwen2-VL: Enhancing Vision-Language Model's Perception of the World at Any Resolution},
  year = 2024,
  journal = {arXiv preprint arXiv:2409.12191},
  author = {Wang, Peng and Bai, Shuai and Tan, Sinan and Wang, Shijie and Fan, Zhihao and Bai, Jinze and Chen, Keqin and Liu, Xuejing and Wang, Jialin and Ge, Wenbin and Fan, Yang and Dang, Kai and others}
}

@inproceedings{wang2024vidproma,
  title = {VidProM: A Million-scale Real Prompt-Gallery Dataset for Text-to-Video Diffusion Models},
  booktitle = {NeurIPS Datasets and Benchmarks Track},
  author = {Wang, Wenhao and Yang, Yi},
  year = 2024
}

@article{wang2025coefficientspreserving,
  title = {Coefficients-Preserving Sampling for Reinforcement Learning with Flow Matching},
  author = {Wang, Feng and Yu, Zihao},
  year = 2025,
  journal = {arXiv preprint arXiv:2509.05952}
}

@article{wang2025grpoguard,
  title = {GRPO-Guard: Mitigating Implicit Over-Optimization in Flow Matching via Regulated Clipping},
  year = 2025,
  journal = {arXiv preprint arXiv:2510.22319},
  author = {Wang, Jing and Liang, Jiajun and Liu, Jie and Liu, Henglin and Liu, Gongye and Zheng, Jun and Pang, Wanyuan and Ma, Ao and Xie, Zhenyu and Wang, Xintao and Wang, Meng and Wan, Pengfei and others}
}

@article{wang2026prefgrpo,
  title = {Pref-GRPO: Pairwise Preference Reward-based GRPO for Stable Text-to-Image Reinforcement Learning},
  author = {Wang, Yibin and Li, Zhimin and Zang, Yuhang and Zhou, Yujie and Bu, Jiazi and Wang, Chunyu and Lu, Qinglin and Jin, Cheng and Wang, Jiaqi},
  year = 2026,
  journal = {arXiv preprint arXiv:2508.20751}
}

@article{wang2026unified,
  title = {Unified Reward Model for Multimodal Understanding and Generation},
  author = {Wang, Yibin and Zang, Yuhang and Li, Hao and Jin, Cheng and Wang, Jiaqi},
  year = 2026,
  journal = {arXiv preprint arXiv:2503.05236}
}

@article{wang2026worldcompass,
  title = {WorldCompass: Reinforcement Learning for Long-Horizon World Models},
  author = {Wang, Zehan and Wang, Tengfei and Zhang, Haiyu and Zuo, Xuhui and Wu, Junta and Wang, Haoyuan and Sun, Wenqiang and Wang, Zhenwei and Cao, Chenjie and Zhao, Hengshuang and Guo, Chunchao and Zhao, Zhou},
  year = 2026,
  journal = {arXiv preprint arXiv:2602.09022}
}

@inproceedings{wangpcflow,
  title = {PC-Flow: Preference Alignment in Flow Matching via Classifier},
  booktitle = {AAAI},
  author = {Wang, Shaomeng and Wang, He and Dai, Longquan and Tang, Jinhui},
  year = 2026
}

@article{wu2025hunyuanvideo,
  title = {HunyuanVideo 1.5 Technical Report},
  year = 2025,
  journal = {arXiv preprint arXiv:2511.18870},
  author = {Wu, Bing and Zou, Chang and Li, Changlin and Huang, Duojun and Yang, Fang and Tan, Hao and Peng, Jack and Wu, Jianbing and Xiong, Jiangfeng and Jiang, Jie and Linus and Patrol and others}
}

@article{wu2025pack,
  title = {Pack and Force Your Memory: Long-form and Consistent Video Generation},
  author = {Wu, Xiaofei and Zhang, Guozhen and Xu, Zhiyong and Zhou, Yuan and Lu, Qinglin and He, Xuming},
  year = 2025,
  journal = {arXiv preprint arXiv:2510.01784}
}

@inproceedings{wu2025rlvrworld,
  title = {RLVR-World: Training World Models with Reinforcement Learning},
  booktitle = {ICLR},
  author = {Wu, Jialong and Yin, Shaofeng and Feng, Ningya and Long, Mingsheng},
  year = 2025
}

@article{xiang2025macrofrommicro,
  title = {Macro-from-Micro Planning for High-Quality and Parallelized Autoregressive Long Video Generation},
  year = 2025,
  journal = {arXiv preprint arXiv:2508.03334},
  author = {Xiang, Xunzhi and Chen, Yabo and Zhang, Guiyu and Wang, Zhongyu and Gao, Zhe and Xiang, Quanming and Shang, Gonghu and Liu, Junqi and Huang, Haibin and Gao, Yang and Zhang, Chi and Fan, Qi and others}
}

@article{xiang2026pathwise,
  title = {Pathwise Test-Time Correction for Autoregressive Long Video Generation},
  author = {Xiang, Xunzhi and Duan, Zixuan and Zhang, Guiyu and Zhang, Haiyu and Gao, Zhe and Wu, Junta and Zhang, Shaofeng and Wang, Tengfei and Fan, Qi and Guo, Chunchao},
  year = 2026,
  journal = {arXiv preprint arXiv:2602.05871}
}

@inproceedings{xu2023imagerewarda,
  title = {ImageReward: Learning and Evaluating Human Preferences for Text-to-Image Generation},
  booktitle = {NeurIPS},
  author = {Xu, Jiazheng and Liu, Xiao and Wu, Yuchen and Tong, Yuxuan and Li, Qinkai and Ding, Ming and Tang, Jie and Dong, Yuxiao},
  year = 2023
}

@inproceedings{yan2024perflow,
  title = {PeRFlow: Piecewise Rectified Flow as Universal Plug-and-Play Accelerator},
  booktitle = {NeurIPS},
  author = {Yan, Hanshu and Liu, Xingchao and Pan, Jiachun and Liew, Jun Hao and Liu, Qiang and Feng, Jiashi},
  year = 2024
}

@inproceedings{yang2024cogvideox,
  title = {CogVideoX: Text-to-Video Diffusion Models with An Expert Transformer},
  booktitle = {ICLR},
  year = 2025,
  author = {Yang, Zhuoyi and Teng, Jiayan and Zheng, Wendi and Ding, Ming and Huang, Shiyu and Xu, Jiazheng and Yang, Yuanming and Hong, Wenyi and Zhang, Xiaohan and Feng, Guanyu and Yin, Da and Gu, Xiaotao and others}
}

@inproceedings{yang2025longlivea,
  title = {LongLive: Real-time Interactive Long Video Generation},
  booktitle = {ICLR},
  author = {Yang, Shuai and Huang, Wei and Chu, Ruihang and Xiao, Yicheng and Zhao, Yuyang and Wang, Xianbang and Li, Muyang and Xie, Enze and Chen, Yingcong and Lu, Yao and Han, Song and Chen, Yukang},
  year = 2026
}

@article{ye2025dataregularized,
  title = {Data-regularized Reinforcement Learning for Diffusion Models at Scale},
  year = 2025,
  journal = {arXiv preprint arXiv:2512.04332},
  author = {Ye, Haotian and Zheng, Kaiwen and Xu, Jiashu and Li, Puheng and Chen, Huayu and Han, Jiaqi and Liu, Sheng and Zhang, Qinsheng and Mao, Hanzi and Hao, Zekun and Chattopadhyay, Prithvijit and Yang, Dinghao and others}
}

@article{ye2025reinforcement,
  title = {Reinforcement Learning with Inverse Rewards for World Model Post-training},
  author = {Ye, Yang and He, Tianyu and Yang, Shuo and Bian, Jiang},
  year = 2025,
  journal = {arXiv preprint arXiv:2509.23958}
}

@inproceedings{yesiltepe2026infinityrope,
  title = {Infinity-RoPE: Action-Controllable Infinite Video Generation Emerges From Autoregressive Self-Rollout},
  booktitle = {CVPR},
  author = {Yesiltepe, Hidir and Meral, Tuna Han Salih and Akan, Adil Kaan and Oktay, Kaan and Yanardag, Pinar},
  year = 2026
}

@article{yi2025deep,
  title = {Deep Forcing: Training-Free Long Video Generation with Deep Sink and Participative Compression},
  author = {Yi, Jung and Jang, Wooseok and Cho, Paul Hyunbin and Nam, Jisu and Yoon, Heeji and Kim, Seungryong},
  year = 2025,
  journal = {arXiv preprint arXiv:2512.05081}
}

@inproceedings{yin2024improved,
  title = {Improved Distribution Matching Distillation for Fast Image Synthesis},
  booktitle = {NeurIPS},
  author = {Yin, Tianwei and Gharbi, Micha{\"e}l and Park, Taesung and Zhang, Richard and Shechtman, Eli and Durand, Fredo and Freeman, William T.},
  year = 2024
}

@inproceedings{yin2024onestep,
  title = {One-Step Diffusion with Distribution Matching Distillation},
  booktitle = {CVPR},
  author = {Yin, Tianwei and Gharbi, Micha{\"e}l and Zhang, Richard and Shechtman, Eli and Durand, Fr{\'e}do and Freeman, William T. and Park, Taesung},
  year = 2024,
  pages = {6613--6623}
}

@inproceedings{yin2025slowa,
  title = {From Slow Bidirectional to Fast Autoregressive Video Diffusion Models},
  booktitle = {CVPR},
  author = {Yin, Tianwei and Zhang, Qiang and Zhang, Richard and Freeman, William T. and Durand, Fredo and Shechtman, Eli and Huang, Xun},
  year = 2025
}

@article{yu2025videossm,
  title = {VideoSSM: Autoregressive Long Video Generation with Hybrid State-Space Memory},
  author = {Yu, Yifei and Wu, Xiaoshan and Hu, Xinting and Hu, Tao and Sun, Yangtian and Lyu, Xiaoyang and Wang, Bo and Ma, Lin and Ma, Yuewen and Wang, Zhongrui and Qi, Xiaojuan},
  year = 2025,
  journal = {arXiv preprint arXiv:2512.04519}
}

@article{yuan2026helios,
  title = {Helios: Real Real-Time Long Video Generation Model},
  author = {Yuan, Shenghai and Yin, Yuanyang and Li, Zongjian and Huang, Xinwei and Yang, Xiao and Yuan, Li},
  year = 2026,
  journal = {arXiv preprint arXiv:2603.04379}
}

@article{yue2026know,
  title = {Know Your Step: Faster and Better Alignment for Flow Matching Models via Step-aware Advantages},
  author = {Yue, Zhixiong and Ni, Zixuan and Ye, Feiyang and Zhang, Jinshan and Shen, Sheng and Mi, Zhenpeng},
  year = 2026,
  journal = {arXiv preprint arXiv:2602.01591}
}

@article{zhang2025blockvid,
  title = {BlockVid: Block Diffusion for High-Quality and Consistent Minute-Long Video Generation},
  author = {Zhang, Zeyu and Chang, Shuning and He, Yuanyu and Han, Yizeng and Tang, Jiasheng and Wang, Fan and Zhuang, Bohan},
  year = 2025,
  journal = {arXiv preprint arXiv:2511.22973}
}

@article{zhang2026astrolabe,
  title = {Astrolabe: Steering Forward-Process Reinforcement Learning for Distilled Autoregressive Video Models},
  author = {Zhang, Songchun and Xue, Zeyue and Fu, Siming and Huang, Jie and Kong, Xianghao and Ma, Y. and Huang, Haoyang and Duan, Nan and Rao, Anyi},
  year = 2026,
  journal = {arXiv preprint arXiv:2603.17051}
}

@inproceedings{zhang2026egrpo,
  title = {E-GRPO: High Entropy Steps Drive Effective Reinforcement Learning for Flow Models},
  booktitle = {CVPR Findings},
  author = {Zhang, Shengjun and Zhang, Zhang and Dai, Chensheng and Duan, Yueqi},
  year = 2026
}

@article{zhang2026opgrpo,
  title = {OP-GRPO: Efficient Off-Policy GRPO for Flow-Matching Models},
  author = {Zhang, Liyu and Li, Kehan and Han, Tingrui and Zhao, Tao and Sheng, Yuxuan and He, Shibo and Li, Chao},
  year = 2026,
  journal = {arXiv preprint arXiv:2604.04142}
}

@article{zhang2026pretraining,
  title = {Pretraining Frame Preservation for Lightweight Autoregressive Video History Embedding},
  author = {Zhang, Lvmin and Cai, Shengqu and Li, Muyang and Zeng, Chong and Lu, Beijia and Rao, Anyi and Han, Song and Wetzstein, Gordon and Agrawala, Maneesh},
  year = 2026,
  journal = {arXiv preprint arXiv:2512.23851}
}

@inproceedings{zhao2026realtime,
  title = {Real-Time Motion-Controllable Autoregressive Video Diffusion},
  booktitle = {ICLR},
  author = {Zhao, Kesen and Shi, Jiaxin and Zhu, Beier and Zhou, Junbao and Shen, Xiaolong and Zhou, Yuan and Sun, Qianru and Zhang, Hanwang},
  year = 2026
}

@article{zhao2026relax,
  title = {Relax Forcing: Relaxed KV-Memory for Consistent Long Video Generation},
  author = {Zhao, Zengqun and Lu, Yanzuo and Liu, Ziquan and Song, Jifei and Deng, Jiankang and Patras, Ioannis},
  year = 2026,
  journal = {arXiv preprint arXiv:2603.21366}
}

@article{zheng2024trajectory,
  title = {Trajectory Consistency Distillation: Improved Latent Consistency Distillation by Semi-Linear Consistency Function with Trajectory Mapping},
  author = {Zheng, Jianbin and Hu, Minghui and Fan, Zhongyi and Wang, Chaoyue and Ding, Changxing and Tao, Dacheng and Cham, Tat-Jen},
  year = 2024,
  journal = {arXiv preprint arXiv:2402.19159}
}

@inproceedings{zheng2026diffusionnft,
  title = {DiffusionNFT: Online Diffusion Reinforcement with Forward Process},
  booktitle = {ICLR},
  author = {Zheng, Kaiwen and Chen, Huayu and Ye, Haotian and Wang, Haoxiang and Zhang, Qinsheng and Jiang, Kai and Su, Hang and Ermon, Stefano and Zhu, Jun and Liu, Ming-Yu},
  year = 2026
}

@article{zheng2026manifoldaware,
  title = {Manifold-Aware Exploration for Reinforcement Learning in Video Generation},
  author = {Zheng, Mingzhe and Kong, Weijie and Wu, Yue and Jiang, Dengyang and Ma, Yue and He, Xuanhua and Lin, Bin and Gong, Kaixiong and Zhong, Zhao and Bo, Liefeng and Chen, Qifeng and Yang, Harry},
  year = 2026,
  journal = {arXiv preprint arXiv:2603.21872}
}

@inproceedings{zhou2024adversarial,
  title = {Adversarial Score identity Distillation: Rapidly Surpassing the Teacher in One Step},
  booktitle = {ICLR},
  author = {Zhou, Mingyuan and Zheng, Huangjie and Gu, Yi and Wang, Zhendong and Huang, Hai},
  year = 2025
}

@inproceedings{zhou2024score,
  title = {Score identity Distillation: Exponentially Fast Distillation of Pretrained Diffusion Models for One-Step Generation},
  booktitle = {ICML},
  author = {Zhou, Mingyuan and Zheng, Huangjie and Wang, Zhendong and Yin, Mingzhang and Huang, Hai},
  year = 2024
}

@inproceedings{zhou2025finegrained,
  title = {Fine-Grained GRPO for Precise Preference Alignment in Flow Models},
  booktitle = {CVPR},
  author = {Zhou, Yujie and Ling, Pengyang and Bu, Jiazi and Wang, Yibin and Zang, Yuhang and Wang, Jiaqi and Niu, Li and Zhai, Guangtao},
  year = 2026
}

@article{zhu2025memorizeandgenerate,
  title = {Memorize-and-Generate: Towards Long-Term Consistency in Real-Time Video Generation},
  author = {Zhu, Tianrui and Zhang, Shiyi and Sun, Zhirui and Tian, Jingqi and Tang, Yansong},
  year = 2025,
  journal = {arXiv preprint arXiv:2512.18741}
}

@inproceedings{zhu2026causal,
  title = {Causal Forcing: Autoregressive Diffusion Distillation Done Right for High-Quality Real-Time Interactive Video Generation},
  booktitle = {ICML},
  author = {Zhu, Hongzhou and Zhao, Min and He, Guande and Su, Hang and Li, Chongxuan and Zhu, Jun},
  year = 2026
}

@article{zhu2026diffusion,
  title = {Diffusion Reinforcement Learning via Centered Reward Distillation},
  author = {Zhu, Yuanzhi and Wang, Xi and Lathuili{\`e}re, St{\'e}phane and Kalogeiton, Vicky},
  year = 2026,
  journal = {arXiv preprint arXiv:2603.14128}
}

@article{zou2026hiar,
  title = {HiAR: Efficient Autoregressive Long Video Generation via Hierarchical Denoising},
  author = {Zou, Kai and Zheng, Dian and Liu, Hongbo and Hang, Tiankai and Liu, Bin and Yu, Nenghai},
  year = 2026,
  journal = {arXiv preprint arXiv:2603.08703}
}

@inproceedings{esser2024scaling,
  title = {Scaling Rectified Flow Transformers for High-Resolution Image Synthesis},
  booktitle = {ICML},
  year = 2024,
  author = {Esser, Patrick and Kulal, Sumith and Blattmann, Andreas and Entezari, Rahim and M\"uller, Jonas and Saini, Harry and Levi, Yam and Lorenz, Dominik and Sauer, Axel and Boesel, Frederic and Podell, Dustin and Dockhorn, Tim and others}
}

@inproceedings{hu2021lora,
  title = {LoRA: Low-Rank Adaptation of Large Language Models},
  booktitle = {ICLR},
  author = {Hu, Edward J. and Shen, Yelong and Wallis, Phillip and {Allen-Zhu}, Zeyuan and Li, Yuanzhi and Wang, Shean and Wang, Lu and Chen, Weizhu},
  year = 2022
}

@inproceedings{kingma2017adam,
  title = {Adam: A Method for Stochastic Optimization},
  booktitle = {ICLR},
  author = {Kingma, Diederik P. and Ba, Jimmy},
  year = 2015
}

@inproceedings{loshchilov2019decoupled,
  title = {Decoupled Weight Decay Regularization},
  booktitle = {ICLR},
  author = {Loshchilov, Ilya and Hutter, Frank},
  year = 2019
}

@article{mcintosh2024isambard,
  title = {Isambard-ai: a leadership-class supercomputer optimised specifically for artificial intelligence},
  author={McIntosh-Smith, Simon and Alam, Sadaf and Woods, Christopher},
  year={2024},
  journal = {arXiv preprint arXiv:2410.11199}
}
}

%%%%%%%%%%%%%%%%%%%%%%%%%%%%%%%%%%%%%%%%%%%%%%%%%%%%%%%%%%%%
\newpage
\appendix

\section{Algorithm Formulations}
\label{app:algorithms}

\begin{algorithm}[t]
\caption{RAVEN training iteration}
\label{alg:raven}
\begin{algorithmic}[1]
\Require causal student $f_{\theta}$, fake-score critic $f_{\phi}$, bidirectional teacher $f_{\psi}$
\Require text condition $c$, chunk count $T$, noise schedule $(\alpha_n, \sigma_n)$
\Require sampling timesteps $\tau_1 > \cdots > \tau_K = 0$
\Require chunk-wise weighting function $g_{\eta}$, learning rates $\eta_{\theta}$ and $\eta_{\phi}$
\Require iteration index $i$, TTUR ratio $r$ (critic-to-generator update frequency)
\Statex \colorbox{black!8}{\makebox[\dimexpr\linewidth-2\fboxsep][l]{\textbf{Stage 1: Self Rollout} \hfill \textit{no gradient through $f_{\theta}$}}}
\For{$t = 1, \ldots, T$}
    \State sample initial noisy state $\hat{z}_t^{(\tau_1)} \sim \mathcal{N}(0, I)$
    \For{$k = 1, \ldots, K-1$}
        \State predict clean endpoint $\hat{x}_t^{(k)} \leftarrow \mathrm{Endpoint}\bigl(f_{\theta}(\hat{z}_t^{(\tau_k)}, \tau_k, c, h_t)\bigr)$
        \State sample $\epsilon \sim \mathcal{N}(0, I)$
        \State consistency transition $\hat{z}_t^{(\tau_{k+1})} \leftarrow \alpha_{\tau_{k+1}} \hat{x}_t^{(k)} + \sigma_{\tau_{k+1}} \epsilon$ \Comment{Eq.~\eqref{eq:consistency_kernel}}
    \EndFor
    \State final endpoint $\hat{x}_t \leftarrow \mathrm{Endpoint}\bigl(f_{\theta}(\hat{z}_t^{(\tau_K)}, \tau_K, c, h_t)\bigr)$
    \State update KV cache with $\hat{x}_t$ so that $h_{t+1} = \mathcal{H}(\hat{x}_{\le t})$
\EndFor
\Statex \colorbox{black!8}{\makebox[\dimexpr\linewidth-2\fboxsep][l]{\textbf{Stage 2: Fake-Score Step}}}
\State concatenate the rollout endpoints into the full clean sequence $\hat{x} = (\hat{x}_1, \ldots, \hat{x}_T)$
\State sample $n \sim \mathcal{U}(0, 1)$ and $\epsilon \sim \mathcal{N}(0, I)$
\State perturb $z^{(n)} \leftarrow \alpha_{n} \hat{x} + \sigma_{n} \epsilon$
\State bidirectional forward pass of $f_{\phi}$ on $z^{(n)}$
\State update critic $\phi \leftarrow \phi - \eta_{\phi} \nabla_{\phi} \bigl\| f_{\phi}(z^{(n)}, n, c) - \mathrm{Target}(\hat{x}) \bigr\|^2$
\Statex \colorbox{black!8}{\makebox[\dimexpr\linewidth-2\fboxsep][l]{\textbf{Stage 3: Generator Step on $\mathcal{I}_u$} \hfill \textit{runs only when $i \bmod r = 0$}}}
\If{$i \bmod r = 0$}
\State sample sampling level $u \in \{\tau_1, \ldots, \tau_{K-1}\}$
\State sample score level $s \sim \mathcal{U}(0, u)$
\State assemble interleaved sequence $\mathcal{I}_{u}$ from the rollout \Comment{Eq.~\eqref{eq:interleaved}}
\State causal forward of $f_{\theta}$ on $\mathcal{I}_{u}$, yielding $\hat{x}_{\theta}$
\State sample $\epsilon \sim \mathcal{N}(0, I)$
\State consistency transition $\hat{z}^{(s)} \leftarrow \operatorname{sg}\bigl(\alpha_{s} \hat{x}_{\theta} + \sigma_{s} \epsilon\bigr)$ \Comment{Eq.~\eqref{eq:consistency_kernel}}
\State teacher endpoint $\hat{x}_{\psi} \leftarrow \mathrm{Endpoint}(f_{\psi}(\hat{z}^{(s)}, s, c))$
\State critic endpoint $\hat{x}_{\phi} \leftarrow \mathrm{Endpoint}(f_{\phi}(\hat{z}^{(s)}, s, c))$
\State per-element DMD loss $\ell \leftarrow \left\| \hat{x}_{\theta} - \operatorname{sg}\!\left( \hat{x}_{\theta} + \dfrac{\hat{x}_{\psi} - \hat{x}_{\phi}}{\|\hat{x}_{\theta} - \hat{x}_{\psi}\|_1} \right) \right\|^2$
\State split $\ell$ along the chunk axis into per-chunk losses $\ell_t$
\State compute future participation scores $p_t$ and chunk weights $w_t \propto g_{\eta}(p_t)$
\State chunk-weighted DMD loss $\mathcal{L}_{\mathrm{DMD}} \leftarrow \dfrac{\sum_{t} w_t \ell_t}{\sum_{t} w_t m_t}$
\State $\theta \leftarrow \theta - \eta_{\theta} \nabla_{\theta} \mathcal{L}_{\mathrm{DMD}}$
\EndIf
\end{algorithmic}
\end{algorithm}

Algorithm~\ref{alg:raven} formalizes one RAVEN training iteration.
The iteration begins with a self rollout of $f_{\theta}$ under the consistency sampler, producing per-chunk denoising trajectories $\{\hat{z}_t^{(\tau_k)}\}$ together with clean endpoints $\hat{x}_t = \hat{z}_t^{(0)}$.
The fake-score critic $f_{\phi}$ is updated on noised endpoints every iteration, whereas the generator $f_{\theta}$ is updated only once every $r$ iterations on the interleaved sequence $\mathcal{I}_u$, with a reverse-KL score gradient defined by the teacher $f_{\psi}$ and the updated critic $f_{\phi}$.
A single causal forward pass over $\mathcal{I}_u$ then routes gradients from later chunks through the cached history $h_t^{\mathrm{RAVEN}} = \mathcal{H}(\hat{x}_{<t})$ used during extrapolation.

\begin{algorithm}[t]
\caption{CM-GRPO training iteration}
\label{alg:cmgrpo}
\begin{algorithmic}[1]
\Require consistency generator $f_{\theta}$ initialized from a RAVEN checkpoint, reward composition $\{R_m\}$ with weights $\{\lambda_m\}$
\Require text condition $c$, group size $G$, chunk count $T$, noise schedule $(\alpha_n, \sigma_n)$
\Require sampling timesteps $\tau_1 > \cdots > \tau_K = 0$
\Require chunk-wise weighting function $g_{\eta}$, advantage clip $A_{\max}$, learning rate $\eta_{\theta}$
\Require normalization stabilization constant $\varepsilon$
\Statex \colorbox{black!8}{\makebox[\dimexpr\linewidth-2\fboxsep][l]{\textbf{Stage 1: Group Rollouts} \hfill \textit{no gradient through $f_{\theta}$}}}
\For{$i = 1, \ldots, G$}
    \For{$t = 1, \ldots, T$}
        \State sample initial noisy state $\tilde{z}_{i,t}^{(\tau_1)} \sim \mathcal{N}(0, I)$
        \For{$k = 1, \ldots, K-1$}
            \State predict clean endpoint $\hat{x}_{i,t}^{(k)} \leftarrow \mathrm{Endpoint}\bigl(f_{\theta}(\tilde{z}_{i,t}^{(\tau_k)}, \tau_k, c, h_{i,t})\bigr)$
            \State sample $\epsilon \sim \mathcal{N}(0, I)$
            \State consistency transition $\tilde{z}_{i,t}^{(\tau_{k+1})} \leftarrow \alpha_{\tau_{k+1}} \hat{x}_{i,t}^{(k)} + \sigma_{\tau_{k+1}} \epsilon$ \Comment{Eq.~\eqref{eq:consistency_kernel}}
        \EndFor
        \State final endpoint $\hat{x}_{i,t} \leftarrow \mathrm{Endpoint}\bigl(f_{\theta}(\tilde{z}_{i,t}^{(\tau_K)}, \tau_K, c, h_{i,t})\bigr)$
        \State update KV cache so that $h_{i,t+1} = \mathcal{H}(\hat{x}_{i, \le t})$
    \EndFor
    \State trajectory endpoint $\hat{x}^{i} \leftarrow (\hat{x}_{i,1}, \ldots, \hat{x}_{i,T})$
\EndFor
\Statex \colorbox{black!8}{\makebox[\dimexpr\linewidth-2\fboxsep][l]{\textbf{Stage 2: Reward Evaluation and Group-Relative Advantages}}}
\For{$i = 1, \ldots, G$}
    \State evaluate each reward dimension $R_m^{i} \leftarrow R_m(\hat{x}^{i}, c)$
\EndFor
\State per-dimension group normalization $\bar{R}_m^{i} \leftarrow \bigl(R_m^{i} - \operatorname{mean}(\{R_m^{j}\}_{j=1}^{G})\bigr) \,/\, \bigl(\operatorname{std}(\{R_m^{j}\}_{j=1}^{G}) + \varepsilon\bigr)$
\State composite reward $R^{i} \leftarrow \sum_{m} \lambda_m \bar{R}_m^{i} \,/\, \sum_{m} |\lambda_m|$
\State group-relative advantage $\hat{A}_i \leftarrow \bigl(R^{i} - \operatorname{mean}(\{R^{j}\}_{j=1}^{G})\bigr) \,/\, \bigl(\operatorname{std}(\{R^{j}\}_{j=1}^{G}) + \varepsilon\bigr)$
\State clip advantage $\hat{A}_i \leftarrow \operatorname{clip}(\hat{A}_i, -A_{\max}, A_{\max})$
\Statex \colorbox{black!8}{\makebox[\dimexpr\linewidth-2\fboxsep][l]{\textbf{Stage 3: Policy Update on Sampled Transitions}}}
\For{$i = 1, \ldots, G$}
    \State sample $k_i \sim \mathcal{U}\{1, \ldots, K-1\}$ and let $u_i \leftarrow \tau_{k_i}$, $s_i \leftarrow \tau_{k_i+1}$
    \State retrieve $\tilde{z}_i^{(u_i)}, \tilde{z}_i^{(s_i)}$ from the cached trajectory
\EndFor
\State predict clean endpoint $\hat{x}_{\theta}^{i} \leftarrow \mathrm{Endpoint}\bigl(f_{\theta}(\tilde{z}_i^{(u_i)}, u_i, c)\bigr)$ for all $i$
\State consistency kernel mean $\mu_{\theta}^{u_i \to s_i} \leftarrow \alpha_{s_i} \hat{x}_{\theta}^{i}$ \Comment{Eq.~\eqref{eq:consistency_kernel}}
\State per-trajectory stop-gradient regression loss
\Statex \hspace{2.6em} $\ell^{i} \leftarrow \left\| \hat{x}_{\theta}^{i} - \operatorname{sg}\!\left( \hat{x}_{\theta}^{i} + \dfrac{\hat{A}_i\, \alpha_{s_i}}{2\sigma_{s_i}^2} \bigl( \tilde{z}_i^{(s_i)} - \mu_{\theta}^{u_i \to s_i} \bigr) \right) \right\|^2$ \Comment{Eq.~\eqref{eq:cmgrpo_loss}}
\State split each $\ell^{i}$ along the chunk axis into per-chunk losses $\ell^{i}_t$
\State compute future participation scores $p_t$ and chunk weights $w_t \propto g_{\eta}(p_t)$
\State chunk-weighted CM-GRPO loss $\mathcal{L}_{\mathrm{CM\text{-}GRPO}} \leftarrow \dfrac{\sum_{i,t} w_t \ell^{i}_t}{\sum_{i,t} w_t m_t}$
\State $\theta \leftarrow \theta - \eta_{\theta} \nabla_{\theta} \mathcal{L}_{\mathrm{CM\text{-}GRPO}}$
\end{algorithmic}
\end{algorithm}

Building on the RAVEN checkpoint, Algorithm~\ref{alg:cmgrpo} formalizes one CM-GRPO training iteration.
The iteration draws a group of $G$ independent consistency rollouts under a shared text condition $c$, scores each rollout with the composite reward, and converts the endpoint scores into a group-relative advantage following Eq.~\eqref{eq:cmgrpo_loss}.
A single transition $u \to s$ is then sampled along each rollout, and the policy update is realized as the stop-gradient regression on the predicted clean endpoint of the consistency kernel, inheriting the chunk-wise weighting introduced by RAVEN so that supervision remains aligned with the autoregressive horizon.
The transition index is restricted to $k_i \in \{1, \ldots, K-1\}$ rather than the full set of consistency steps, since the final transition lands at $\tau_K = 0$ with $\sigma_{\tau_K} = 0$, where the kernel collapses to a Dirac delta and the policy log-probability is undefined.

\section{More Implementation Details}
\label{app:implementation_details}

\noindent\textbf{Dataset.}
Both RAVEN and CM-GRPO are trained exclusively on text prompts drawn from VidProM~\cite{wang2024vidproma}, preprocessed through filtering and large language model extension, following the data protocol of Self Forcing~\cite{huang2025selfa}.
Ablation experiments that require real video data draw from OpenVidHD-0.4M~\cite{nan2025openvid1ma}, with all video clips temporally upsampled via RIFE~\cite{huang2022realtime} prior to use.

\noindent\textbf{Training Details.}
Most RAVEN training settings are inherited from Self Forcing~\cite{huang2025selfa} and Causal Forcing~\cite{zhu2026causal}.
We disable weight decay throughout and reduce the two-time-scale update rule (TTUR)~\cite{yin2024improved} ratio between critic and generator updates from $5$ to $2$.
CM-GRPO is instead optimized in a parameter-efficient manner on top of the RAVEN checkpoint, applying LoRA~\cite{hu2021lora} of rank $256$ to all linear layers under AdamW~\cite{kingma2017adam,loshchilov2019decoupled} with learning rate $5\times 10^{-6}$, zero weight decay, betas $(0.0, 0.999)$ and epsilon $10^{-10}$, where each policy update consumes a batch size of $8$ paired with a group size of $32$.
The training-time test framework and chunk-wise loss scaling introduced by RAVEN are carried over unchanged into the CM-GRPO stage, so that the policy update inherits the same alignment between training context and inference-time extrapolation.
The Causal Forcing~\cite{zhu2026causal} + CM-GRPO entry in Table~\ref{tab:quantitative_results} does not utilize these two designs, so as to isolate the contribution of the policy objective from RAVEN itself.
For this entry, the learning rate is lowered to $2\times 10^{-6}$ and weight decay is reinstated at $0.01$, while the reward weight on the \textit{Dynamic Degree} (DD) dimension is raised from $0.35$ to $2.35$ to compensate for the weaker motion of the Causal Forcing checkpoint.
RAVEN and CM-GRPO consume training budgets of approximately $70$ and $170$ NVIDIA H200 GPU hours respectively.

\noindent\textbf{Reward Composition.}
Temporal dynamics are captured through dynamic degree and motion smoothness, where the former takes the top-5\% mean of optical flow magnitudes estimated by RAFT~\cite{teed2020raft} across consecutive frame pairs as an index of peak scene motion, and the latter is derived from the reconstruction error incurred when AMT~\cite{li2023amt} recovers artificially dropped frames from their temporal neighbors, with lower error reflecting greater temporal coherence.
At the frame level, the LAION aesthetic predictor~\cite{laion-ai2022aestheticpredictor} applies a linear model on image embeddings to score each frame for compositional appeal and perceptual naturalness, while the multi-scale MUSIQ~\cite{ke2021musiq} evaluates low-level technical distortion.
Text-video alignment is additionally assessed by VideoReward~\cite{liu2025improving}, a reward model built on Qwen2-VL~\cite{wang2024qwen2vl} and trained via direct preference optimization~\cite{rafailov2024direct} on 182K pairwise human preference annotations.

\section{User Study}
\label{app:user_study}

\begin{figure}[t]
    \centering
    \includegraphics[width=\linewidth]{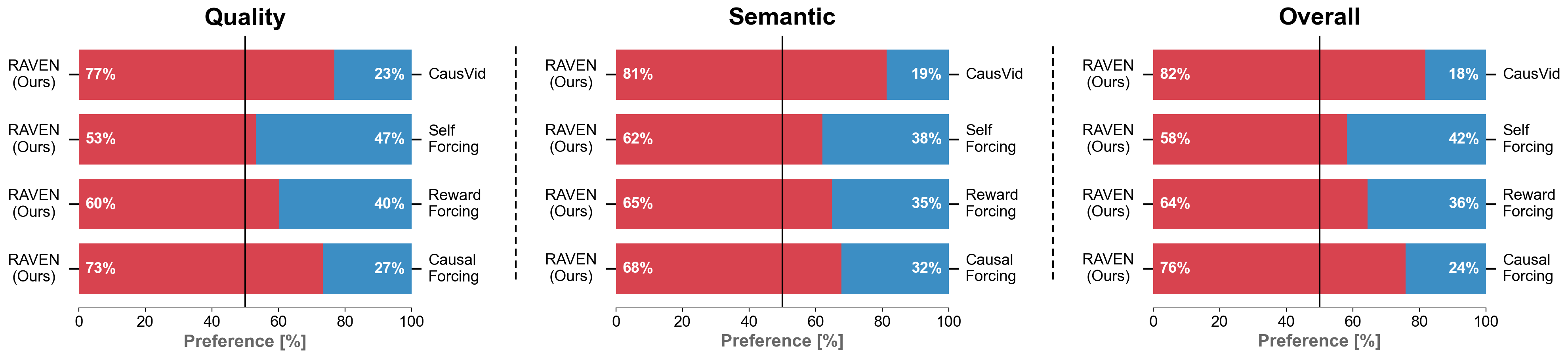}
    \caption{User study preference rates on \textit{Quality}, \textit{Semantic}, and \textit{Overall}.}
    \label{fig:preference_combined}
\end{figure}

\begin{figure}[t]
    \centering
    \includegraphics[width=\linewidth]{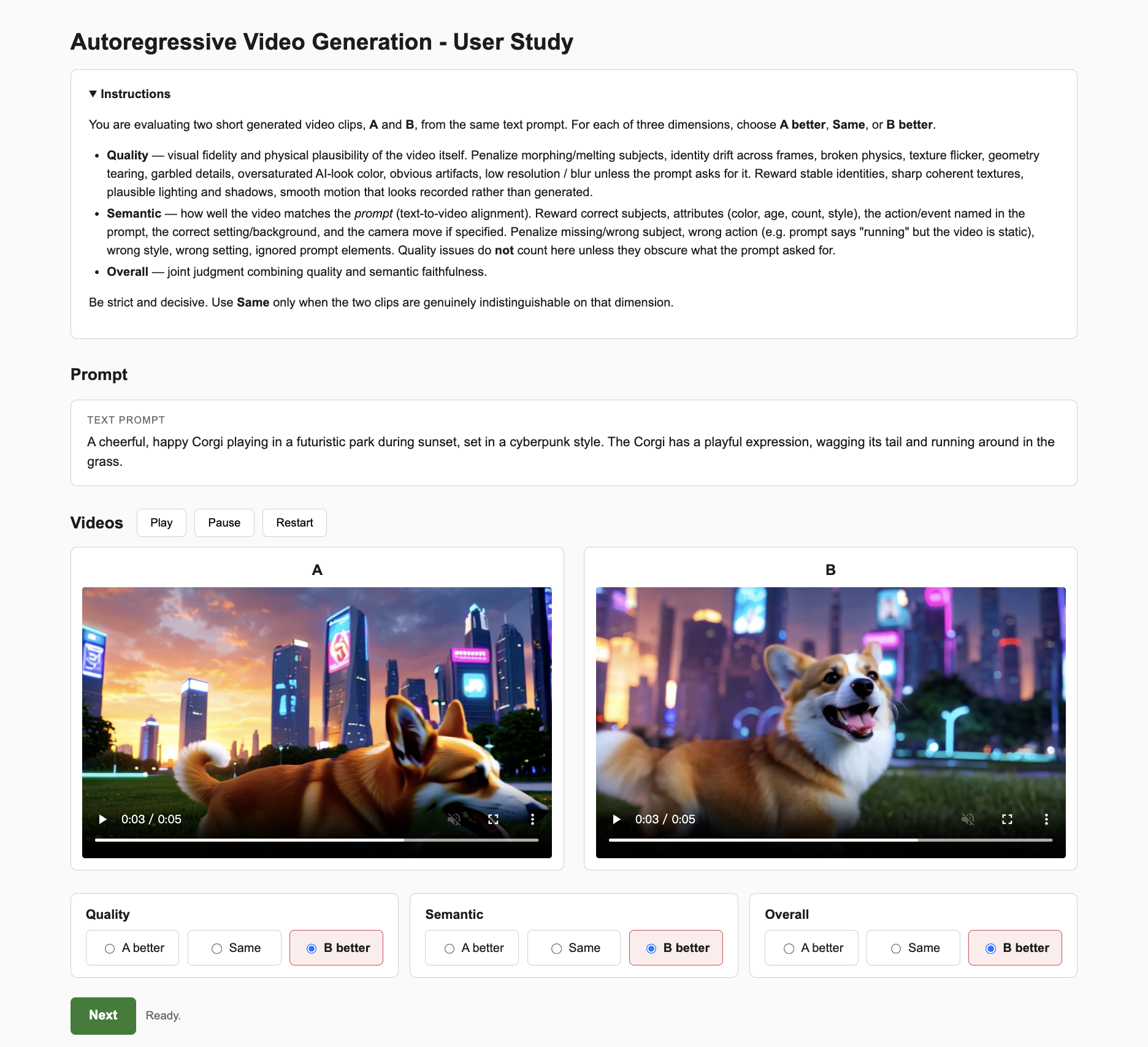}
    \vspace{-10pt}
    \caption{User study instruction screenshot.}
    \vspace{-5pt}
    \label{fig:user_study_page}
\end{figure}

We conduct a user study on $100$ long and detailed prompts drawn from the qualitative showcases of the existing baselines, generating $4$ samples per prompt for each method.
The study covers the four baselines designed for $5$-second short video generation, namely CausVid~\cite{yin2025slowa}, Self Forcing~\cite{huang2025selfa}, Reward Forcing~\cite{lu2025reward}, and Causal Forcing~\cite{zhu2026causal}.
For each sample pair, an individual user rates a RAVEN clip against its baseline counterpart presented in randomized order along \textit{Quality}, \textit{Semantic}, and \textit{Overall}, following the instructions in Figure~\ref{fig:user_study_page}.
Aggregate preference rates are reported in Figure~\ref{fig:preference_combined}, where RAVEN is preferred on every dimension against all four baselines, with a more pronounced lead on \textit{Semantic} than on \textit{Quality} and a clear margin on \textit{Overall}.

\section{Discussion}
\label{app:discussion}

Although RAVEN and CM-GRPO are presented with concrete design choices tailored to causal autoregressive video distillation, both formulations admit broader scope than the setting evaluated in our experiments.
The interleaved sequence construction underlying RAVEN currently treats clean chunks as historical context, yet the supervised forward pass does not restrict the form of cached history.
Arbitrary representations derived from the rollout can be substituted in its place, including intermediate noisy states from preceding chunks, memory tokens specific to the underlying architecture, and cache management strategies such as sliding windows or attention sinks, provided the same representation is consumed by subsequent chunks during self rollout.
These representations can in principle be instantiated inside the supervised forward pass and optimized end-to-end through downstream losses on later noisy states, which would turn history handling into a learnable rather than fixed element of the rollout.

CM-GRPO admits an analogous extension along a different axis.
Its policy interface depends only on the conditional Gaussian transition induced by the consistency sampler and is independent of the autoregressive structure of the generator on which it is applied.
The same objective therefore applies to any few-step generator that draws samples through a stochastic consistency step, encompassing bidirectional video models as well as generators in other modalities that have been distilled through consistency training or distillation.
Collectively, RAVEN constitutes a general training-time test interface for autoregressive few-step distillation, while CM-GRPO constitutes a general policy optimization interface for consistency generators, indicating that both contributions extend beyond the specific setting studied here.

\section{Prompts}
\label{app:prompts}

We list the text prompts behind the qualitative comparisons in the main text, indexed by their position within each figure grid.

\noindent\textbf{Figure~\ref{fig:qualitative_comparison} (qualitative comparison).}
\begin{itemize}[leftmargin=1.5em]
\item \textit{Top-left.} A cheerful, happy Corgi playing in a futuristic park during sunset, set in a cyberpunk style. The Corgi has a playful expression, wagging its tail and running around in the grass. The park is illuminated by neon lights and surrounded by towering skyscrapers with holographic advertisements flashing on their surfaces. The sky is a blend of orange and purple hues, creating a striking contrast against the dark cityscape. The Corgi is in the foreground, while the vibrant city lights and buildings create a dynamic background. The scene captures the essence of a cyberpunk world with natural elements intertwined. Medium close-up shot focusing on the Corgi's joyful playfulness.
\item \textit{Top-right.} A majestic African lion with golden fur and piercing green eyes grabs a small antelope tightly in its powerful jaws. The lion's mane flows gracefully as it crouches on all fours, muscles tense and alert. After a moment of contemplation, the lion releases the antelope, which quickly runs away. The scene takes place in a dense savanna with tall grass and scattered acacia trees. The lion remains still for a few moments before turning its gaze back to the fleeing antelope. Wide shot, with a focus on the lion's expressions and movements.
\item \textit{Bottom-left.} A cheerful woman with a warm smile, standing outdoors under a clear blue sky. Her hair is flowing freely as she faces the wind, creating a gentle, swirling wind effect around her. She is wearing a light, flowy dress that moves gracefully with the breeze. The background includes green trees and blooming flowers, adding a vibrant and lively atmosphere. The scene captures a close-up of the woman, emphasizing her joyful expression and the dynamic wind movement around her.
\item \textit{Bottom-right.} On a rainy day, a young boy with tousled brown hair is sprinting away from the rain, his clothes slightly damp and clinging to his body. His face is animated with excitement and joy as he runs barefoot through puddles, splashing water everywhere. The sky is overcast, with heavy raindrops falling steadily. In the background, blurred figures of people walking with umbrellas can be seen. The scene is captured in a medium shot, focusing on the boy's dynamic movements and expressions.
\end{itemize}

\noindent\textbf{Figure~\ref{fig:ablation_qualitative} (qualitative ablation).}
\begin{itemize}[leftmargin=1.5em]
\item \textit{Left.} A documentary-style nature photography shot from a camera truck moving to the left, capturing a crab quickly scurrying into its burrow. The crab has a hard, greenish-brown shell and long claws, moving with determined speed across the sandy ground. Its body is slightly arched as it burrows into the sand, leaving a small trail behind. The background shows a shallow beach with scattered rocks and seashells, and the horizon features a gentle curve of the coastline. The photo has a natural and realistic texture, emphasizing the crab's natural movement and the texture of the sand. A close-up shot from a slightly elevated angle.
\item \textit{Middle.} A dynamic action shot of a surfer accelerating on a powerful wave, carving through the water with grace and agility. The surfer, with a tanned complexion and muscular build, rides the wave with one hand gripping the board while the other extends outwards for balance. The water splashes behind, creating a foamy trail, and the sun casts a golden glow over the scene. The background features a clear blue ocean and distant white-capped waves, with a few seagulls flying overhead. The surfer's expression is one of exhilaration and focus. A mid-shot from a low-angle perspective capturing the surfer's motion and the wave's power.
\item \textit{Right.} A scenic photograph capturing the moment a steam train departs from the Glenfinnan Viaduct, a historic railway bridge in Scotland. The train moves gracefully over the arch-covered viaduct, its smoke billowing into the air. The landscape is lush with greenery, and towering rocky mountains frame the scene, creating a picturesque backdrop. The sky is a clear, bright blue with the sun shining down, casting a warm glow on the train and the surrounding scenery. The viaduct itself is a striking feature, with intricate ironwork and a verdant setting. The photo has a classic, nostalgic feel, emphasizing the natural beauty and historical charm of the location. A wide-angle shot from a slightly elevated angle, capturing both the train and the expansive landscape.
\end{itemize}

\section{Broader Impacts}
\label{app:broader_impacts}

RAVEN and CM-GRPO advance the practical viability of real-time autoregressive video generation, which carries both positive and negative societal implications.
On the positive side, real-time low-latency video synthesis can support creative tools, education, accessibility applications, and interactive simulation, while reducing the compute footprint of video generation pipelines and lowering the barrier for downstream research.
On the negative side, more capable and more accessible video generators raise familiar misuse concerns, including the production of misleading or non-consensual synthetic media, identity impersonation, and content that could be used to spread misinformation.
These risks are inherited from the underlying generative models rather than introduced by our contributions, but the gain in efficiency does broaden the population of users who can produce such content.
We encourage downstream applications to combine model release with provenance tooling, watermarking, and content moderation, and to follow community norms for the responsible release of generative video models.

%%%%%%%%%%%%%%%%%%%%%%%%%%%%%%%%%%%%%%%%%%%%%%%%%%%%%%%%%%%%

% \newpage
% \input{checklist.tex}

\end{document}